\PassOptionsToPackage{table}{xcolor}
\documentclass[9pt,a4paper]{rho-class/rho}

\usepackage[backend=biber, style=nature]{biblatex}
\usepackage[english]{babel}
\usepackage{multirow}
\usepackage{booktabs}
\usepackage{pdfpages}
\usepackage{graphicx}
\usepackage{tabularx}
\usepackage{caption}
\usepackage{subcaption}

\definecolor{LightGrey}{HTML}{F0F4F7}
\definecolor{LightBlue}{HTML}{D6EAF8}
\definecolor{rhocolor}{rgb}{0.0,0.2,0.4} 

\setbool{rho-abstract}{true}
\setbool{corres-info}{true}

\title{Uncertainty Modeling in Multimodal Speech Analysis Across the Psychosis Spectrum}

\author[1]{Morteza Rohanian}
\author[2]{Roya M. Hüppi}
\author[1]{Farhad Nooralahzadeh}
\author[2]{Noemi Dannecker}
\author[2]{Yves Pauli}
\author[2]{Werner Surbeck}
\author[3]{Iris Sommer}
\author[4,5]{Wolfram Hinzen}
\author[6]{Nicolas Langer}
\author[2]{Michael Krauthammer}
\author[2,7]{Philipp Homan}

\affil[1]{Department of Quantitative Biomedicine, University of Zurich}
\affil[2]{Department of Adult Psychiatry and Psychotherapy, University of Zurich}
\affil[3]{Department of Neuroscience, University Medical Center Groningen, Antoni Deusinglaan 2, room 117 Groningen, Netherland}
\affil[4]{Department of Translation and Language Sciences, Universitat Pompeu Fabra, Barcelona, Spain}
\affil[5]{Institució Catalana de Recerca i Estudis Avançats (ICREA), Barcelona, Spain}
\affil[6]{Department of Psychology, University of Zurich}
\affil[7]{Neuroscience Center Zurich, University of Zurich and ETH Zurich, Zurich, Switzerland}

\leadauthor{Rohanian et al.}
\footinfo{Creative Commons CC BY 4.0}
\license{Creative Commons CC BY 4.0}
\smalltitle{Uncertainty modeling in psychotic language}
\institution{}
\theday{February 2025}

\corres{Philipp Homan, MD, PhD}
\email{philipp.homan@bli.uzh.ch}
\begin{abstract}
Capturing subtle speech disruptions across the psychosis spectrum is challenging because of the inherent variability in speech patterns. This variability reflects individual differences and the fluctuating nature of symptoms in both clinical and non-clinical populations. Accounting for uncertainty in speech data is essential for predicting symptom severity and improving diagnostic precision. Speech disruptions characteristic of psychosis appear across the spectrum, including in non-clinical individuals. We develop an uncertainty-aware model integrating acoustic and linguistic features to predict symptom severity and psychosis-related traits. Quantifying uncertainty in specific modalities allows the model to address speech variability, improving prediction accuracy. We analyzed speech data from 114 participants, including 32 individuals with early psychosis and 82 with low or high schizotypy, collected through structured interviews, semi-structured autobiographical tasks, and narrative-driven interactions in German. The model improved prediction accuracy, reducing RMSE and achieving an F1-score of 83\% with ECE = 4.5e-2, showing robust performance across different interaction contexts. Uncertainty estimation improved model interpretability by identifying reliability differences in speech markers such as pitch variability, fluency disruptions, and spectral instability. The model dynamically adjusted to task structures, weighting acoustic features more in structured settings and linguistic features in unstructured contexts. This approach strengthens early detection, personalized assessment, and clinical decision-making in psychosis-spectrum research.
\end{abstract}

\keywords{Schizophrenia spectrum; language and speech analysis; multimodal model; uncertainty modeling; schizotypy}

\begin{document}
\maketitle
\thispagestyle{firststyle}    


\section{Introduction}

Psychosis is a complex condition that disrupts regular mental and social functioning, affecting cognition, perception, emotional stability, and interpersonal relationships. Reflecting the variability in the neurobiology of the disorder \cite{Omlor2025}, symptoms such as hallucinations, delusions, social withdrawal, and cognitive impairment vary between individuals and over time, leading to heterogeneous treatment results with frequent relapses \cite{insel2010rethinking, keeley2018symptom, Winkelbeiner2019,Homan2022f}. Differences in cognitive, emotional, and linguistic disruptions appear across individuals, while symptom severity and presentation can fluctuate within the same person over time. These variations complicate assessments and highlight the need for objective, reproducible measures that capture psychosis-related traits more reliably \cite{griswold2015recognition, phillips2020rethinking}.

While clinical interviews and self-report questionnaires are used to diagnose a psychiatric disorder, such diagnoses have no clear biological correlates \cite{sellbom2020cambridge}. In fact, diagnostic categories may overlook the full complexity of psychiatric disorders and may compromise personalized treatment plans \cite{kvig2023does,Palaniyappan2022,Corcoran2020,Corcoran2020a}. To address these limitations, quantifiable and reproducible measures are required that can better quantify symptom variability and may ultimately guide clinical decision making. Analyzing speech patterns offers a promising behavioral signal method to detect subtle changes in cognitive and emotional states \cite{de2023acoustic, dikaios2023applications, He2024,Corona2022,Palominos2024} and may also be used to monitor therapeutic interventions \cite{Ben-Zion2025,Panchalingam2025}.

Speech analysis captures disturbances in psychosis because key symptoms, such as tangentiality, derailment, and flat affect, frequently manifest in speech patterns. Acoustic features (e.g., pitch variability, spectral changes) reflect vocal expressivity and motor control, while linguistic features (e.g., word choice, coherence) capture disturbances in thought organization \cite{voppel2023semantic,Parola2022}. Monotone speech and reduced prosodic variability often correlate with negative symptoms, such as flattened affect, indicating diminished emotional and social engagement \cite{hitczenko2021understanding, de2023acoustic, dikaios2023applications,Parola2020}.

Recognizing that psychosis traits appear even in the general population (known as schizotypal traits) is crucial to understanding the psychosis spectrum \cite{Sarti2025,Kirchhoff2024}. Schizotypy includes a variety of personality traits and experiences, including unusual perceptual experiences, odd beliefs, and social difficulties, that mirror but are less severe than those of psychosis. Individuals with high levels of schizotypal traits can show subtle speech disturbances, reflecting underlying cognitive and perceptual differences, even without a clinical diagnosis \cite{kiang2010schizotypy, minor2010affective, cohen2014normalities}. Identifying speech markers along this continuum provides insights into underlying psychopathology and supports the dimensional approach suggesting that schizotypal traits exist along a continuum with psychosis \cite{mason2015assessment}.

Identifying subtle speech disruptions across the psychosis spectrum is challenging due to the inherent variability in speech patterns, as speech-based models are influenced by data inconsistencies, contextual factors, and biases in feature selection and interpretation \cite{gawlikowski2023survey}. This variability manifests in different ways, including disruptions in language connectedness, which distinguish schizophrenia symptoms from other conditions and emphasize structural alterations in speech organization \cite{voppel2021quantified,He2024,Palominos2024}. Differences in vocal expressivity further complicate identification, as some individuals maintain normal prosody, challenging assumptions about monotone speech as a consistent marker of negative symptoms \cite{cohen2016vocal}. Monotone speech inconsistencies across clinical assessments reveal limitations in subjective evaluations \cite{compton2018aprosody}. This variability not only arises from individual differences in speech production but also reflects the dynamic nature of symptom presentation in both clinical and non-clinical populations. Machine learning models must account for variability in speech abnormalities to prevent misclassification and ensure robust performance across diverse populations \cite{bone2017signal}. External validation remains essential to integrating speech-based models into clinical practice, minimizing bias and improving diagnostic reliability \cite{chekroud2021promise}.  Accounting for uncertainty in speech data is essential to reliably predict symptom severity and improve diagnostic precision. Informing clinicians about model uncertainty enhances trust and supports better treatment decisions \cite{van2023psychosis}.  
In addition, the impact of different interaction contexts on the effectiveness of speech-based models remains underexplored. There is a need for research that integrates uncertainty-aware modeling with multimodal speech analysis across various interaction settings.  Recent advancements in machine learning, particularly in uncertainty-aware modeling, enable us to analyze speech data more thoroughly while accounting for inherent variability \cite{mcknight2023uncertainty,schrufer2024you,  dighe2024leveraging}.

We distinguish two primary types of uncertainty that impact speech-based psychosis research: aleatoric (data) uncertainty and epistemic (model) uncertainty \cite{kendall2017uncertainties}. Aleatoric uncertainty arises from variability or noise in speech data: background sounds, fluctuating recording conditions, or transient changes in how a person expresses symptoms. Epistemic uncertainty stems from a model’s lack of knowledge; insufficient training examples or unrepresented symptom subtypes lead to higher uncertainty. Researchers in mental health have increasingly recognized the importance of quantifying both kinds of uncertainty, since modeling them can improve symptom prediction and enhance trust in clinical applications \cite{kompa2021second,popat2023embracing,kang2024cure}. When the acoustic modality is noisy and has high aleatoric uncertainty, our models reduce its influence. High epistemic uncertainty, on the other hand, indicates limited experience with a particular symptom profile or patient group. Accounting for both types of uncertainty in multimodal analysis enhances confidence and adaptability in predictions.

Our primary objective was to examine the relationships between acoustic and semantic speech features and positive, negative, and disorganized symptoms across the psychosis spectrum. We explored how different interaction contexts (such as monologue versus dialogue, conversational versus narrative, and spontaneous versus scripted settings) influence the models' ability to predict schizotypal traits and symptoms from speech patterns. Speech patterns can vary significantly depending on the context in which they are observed. For example, monologues may reveal more disorganized thought processes, while dialogues can provide insights into social engagement and cognitive flexibility. Structured tasks may elicit more direct responses related to symptom severity, whereas unstructured sessions might highlight natural speech patterns \cite{de2020artificial}.
To this end, we developed a multimodal model that integrates text and audio using Temporal Context Fusion (TCF). This approach dynamically adjusts the contribution of acoustic and linguistic inputs based on signal quality, improving robustness against noise and inconsistencies in speech data. As we will show, modeling data reliability at both modality and context levels refined the detection of psychosis-related traits and ensured adaptability across different interaction settings.

\section{Materials and Experimental Procedures}

This study examines the relationship between acoustic and semantic speech features and psychosis-spectrum traits across different interaction contexts. To achieve this, we collected multimodal speech data from clinical and non-clinical participants, incorporating structured and semi-structured tasks. Speech-based machine learning techniques were applied to analyze patterns linked to cognitive and perceptual differences across the spectrum.

\subsection{Participants and Study Procedures}

This study was conducted as part of the cross-sectional research project titled "Ventral language stream in schizophrenia with regard to semantic and visuo-spatial processing anomalies (VELAS)" \cite{Sarti2025}. A total of \textit{n} = 114 individuals participated, including \textit{n} = 32 individuals with early psychosis and \textit{n} = 82 individuals with low or high schizotypy. Patients met the International Statistical Classification of Diseases and Related Health Problems 10th Revision (ICD-10) criteria for various psychotic disorders, specifically schizophrenia (F20; \textit{n} = 18), acute and transient psychotic disorder (F23; \textit{n} = 6), schizoaffective disorder (F25; \textit{n} = 5), and recurrent depressive disorder with severe psychotic symptoms (F33.3; \textit{n} = 3). To be included in the study, patients needed to be within the first eight years of psychosis onset, admitted to the University Hospital of Psychiatry in Zurich, Switzerland, and aged 14 to 40 years. Table~\ref{tab:demographics} summarizes the demographic and clinical characteristics of the study participants across schizotypy and patient groups.

Healthy individuals with low or high schizotypy were recruited online: 1061 individuals were initially screened online for schizotypy using the Multidimensional Schizotypy Scale (MSS) \cite{mason1995new, mason2006oxford} and the Oxford-Liverpool Inventory of Feelings and Experiences (O-LIFE) \cite{kwapil2018development}. The same questionnaires, MSS and O-LIFE, were administered to assess schizotypy among the patients with psychosis. A cluster analysis, including the scores of the psychosis group, identified five (MSS) and four (O-LIFE) clusters within the respective schizotypy questionnaires. As the patients were found to be in clusters 3-5 (MSS) or 2-4 (O-LIFE) on the basis of their responses, the people in the low schizotypy group were selected from clusters 1-2 (MSS) or 1 (O-LIFE). The high schizotypy group was selected from individuals with MSS and O-LIFE scores that clustered with the patient group. Both groups were selected to match the patient group in terms of gender, educational level and age. Finally, \textit{n} = 45 individuals were assigned to the low schizotypy group and \textit{n} = 37 to the high schizotypy group. An exclusion criterion for healthy individuals was a history of any psychiatric disorder. All participants had to be right-handed, as determined by the Edinburgh Handedness Inventory short form \cite{veale2014edinburgh}, and abstinent from substance intoxication or withdrawal, including cannabis use, for at least one month. Further exclusion criteria entailed ophthalmological or neurological conditions, head trauma, any prior episodes of lack of consciousness, and pregnancy. All participants gave written informed consent.

Each participant underwent a semi-structured autobiographical interview and a photo-elicitation task incorporating elements of the Thematic Apperception Test (TAT). The Structured Clinical Interview for the Positive and Negative Syndrome Scale (SCI-PANSS) was administered to all subjects to evaluate their psychotic symptom severity on the PANSS \cite{kay1987positive} and its negative, positive, and general subscales. Finally, participants completed a semi-structured interview according to the DISCOURSE in Psychosis consortium protocol, which included three TAT images not used in the photo-elicitation task\footnote{For more information, see: \url{https://discourseinpsychosis.org/resources/}}.

\begin{table}[t]
\centering
\small  
\rowcolors{2}{LightGrey}{white}  
\begin{tabularx}{\columnwidth}{p{4.5cm}|>{\centering\arraybackslash}X>{\columncolor{LightBlue}\centering\arraybackslash}X>{\columncolor{LightBlue}\centering\arraybackslash}X}
\toprule
\textbf{Variable} & \textbf{Low Schizotypy (n=45)} & \textbf{High Schizotypy (n=37)} & \textbf{Patient (n=32)} \\
\midrule
\textbf{Sex} & & & \\
Female & 27 & 17 & 8 \\
Male & 18 & 20 & 24 \\
\midrule
\textbf{Age (Mean (SD), years)} & 25 (4.1) & 27.2 (4.6) & 26.7 (5.7) \\
\midrule
\textbf{Education} & & & \\
Secondary School & 1 & 0 & 6 \\
Vocational Training & 13 & 10 & 15 \\
Higher Vocational School & 3 & 2 & 0 \\
Baccalaureate ("Matura") & 12 & 9 & 3 \\
University (Bachelor’s, incl. Applied Sciences) & 12 & 11 & 3 \\
University (Master’s/Doctorate) & 4 & 5 & 5 \\
\midrule
\textbf{First Language} & & & \\
German & 36 & 28 & 21 \\
Bilingual & 8 & 7 & 9 \\
Other & 1 & 2 & 2 \\
\midrule
\textbf{Diagnosis} & & & \\
Schizophrenia (F20) & – & – & 18 \\
Acute Psychotic Disorder (F23) & – & – & 6 \\
Schizoaffective Disorder (F25) & – & – & 5 \\
Recurrent Depressive Disorder (F33.3) & – & – & 3 \\
\midrule
\textbf{Age of Onset (Mean (SD), years)} & – & – & 25.3 (7) \\
\midrule
\textbf{Number of Episodes} & & & \\
1 & – & – & 17 \\
2 & – & – & 11 \\
3 & – & – & 3 \\
4 & – & – & 1 \\
\midrule
\textbf{Antipsychotic Therapy} & & & \\
None & 45 & 37 & 6 \\
Monotherapy & – & – & 20 \\
Polytherapy ($\geq$2 Antipsychotics) & – & – & 6 \\
\midrule
\textbf{Duration of Antipsychotic Medication (Mean (SD), weeks)} & – & – & 65 (95) \\
\midrule
\textbf{Duration of Illness (Mean (SD), weeks)} & – & – & 140 (126.4) \\
\bottomrule
\end{tabularx}
\caption{Demographic and clinical characteristics of participants across schizotypy and patient groups.}
\label{tab:demographics}
\end{table}

\subsection{Assessments and Measures}

The Positive and Negative Syndrome Scale (PANSS)  \cite{kay1987positive} is a widely used tool for assessing schizophrenia symptoms. The positive subscale (PANSS-Pos) comprises seven items covering symptoms such as hallucinatory behavior and delusions, while the negative subscale (PANSS-Neg) includes seven items that assess symptoms such as blunted affect, lack of spontaneity and flow of conversation, and difficulty in abstract thinking. These scales are supplemented by a general psychopathology scale with 16 items addressing anxiety, depression, and impulse control, amongst others, creating a comprehensive 30-item assessment.

The Multidimensional Schizotypy Scale (MSS) \cite{kwapil2018development} is a self-report instrument designed to evaluate three schizotypy dimensions: positive (MSS-Pos), negative (MSS-Neg), and disorganized (MSS-Dis). It consists of 77 true/false items. Positive schizotypy items include themes such as supernatural experiences and magical beliefs. The negative dimension measures traits such as social disinterest and flattened affect, while the disorganized scale assesses disruptions and disorganization in thought, speech, and behavior \cite{kwapil2018development}.

The Oxford-Liverpool Inventory of Feelings and Experiences (O-LIFE) \cite{mason1995new} is used to assess proneness to psychosis and therefore schizotypy. It includes four scales across 104 true/false items: unusual experiences, introvertive anhedonia, cognitive disorganization, and impulsive nonconformity. Unusual experiences reflects positive schizotypy traits such as perceptual aberrations and magical thinking. Introvertive anhedonia is related to negative schizotypy, including lack of enjoyment from physical and social sources of pleasure and avoiding intimacy. Cognitive disorganization represents thought disorder and other aspects of disorganization commonly seen in psychotic disorders, such as impaired concentration and social anxiety. Finally, Impulsive nonconformity measures impulsive, antisocial behaviors often indicating limited self-control \cite{mason2006oxford}.

\begin{figure*}[ht!]
    \centering
    \includegraphics[width=0.9\textwidth]{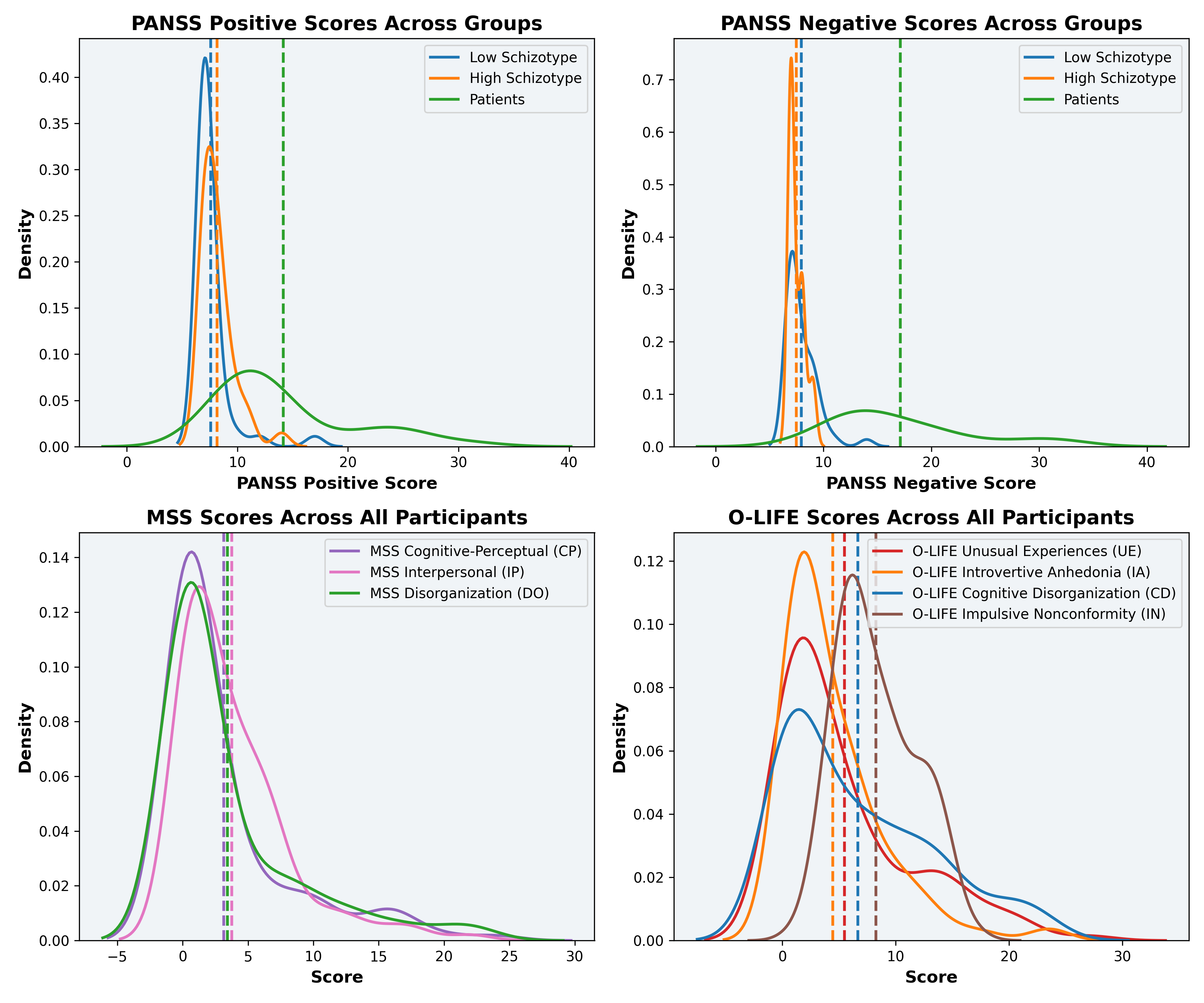}
    \caption{KDE plots with mean lines for different psychological measures. The top two subplots (a, b) represent PANSS Positive and Negative scores across groups (low schizotypy, high schizotypy, and patients). The bottom two subplots (c, d) represent MSS and O-LIFE scores across all participants. The dashed vertical lines indicate the mean values for each group (a, b) or each subscale (c, d). The number of participants in each group is as follows: low schizotypy (\textit{n = 45}), high schizotypy (\textit{n = 37}), and patients (\textit{n = 32}).}
    \label{fig:panss_mss_olife}
\end{figure*}

\section{Methodology}

We developed a model that integrates text and audio to analyze symptoms and schizotypy traits across the psychosis spectrum using uncertainty-aware machine learning and Temporal Context Fusion (TCF) to handle multimodal data complexities.

\subsection{Data Preprocessing}

\textbf{Speaker Diarization} To focus our analysis on patient speech, we performed automatic speaker diarization using the \textit{pyannote.audio} toolkit \cite{bredin2021end}. We enhanced the accuracy of subsequent analyses by isolating the patient's speech and excluding contributions from interviewers and background noise.

\textbf{Segmentation and Alignment} We performed word and phoneme-level segmentation using WhisperX \cite{bain2023whisperx, Spiller2023}, which aligns speech with text transcriptions using wav2vec 2.0 \cite{baevski2020wav2vec} for forced alignment. We ensured that the extracted acoustic features corresponded precisely to specific verbal responses, enabling more granular feature analysis. Since the interviews were conducted in German, we adapted the tools to support the German language.

We aligned the extracted speech data with the structure of each interview session through session-specific segmentation. In structured DISCOURSE sessions, we also segmented the speech at the task level to capture verbal behaviors related to specific prompts. For semi-structured interviews, we use a simple type of topic modeling on interview transcripts, similar to \cite{gong2017topic}. First, we examined all of the interviewer's words to identify a group of unique sentences. This stage created an initial collection, which we then refined by manually deleting sentences that do not introduce new themes. Once the filtered set was complete, we clustered the sentences according to their thematic similarity. The clustering happened in two stages. In the first step, we combined sentences that differ by no more than three characters. In the second step, we manually inspected these clusters to confirm accuracy and make improvements as needed. Finally, we associated each cluster with a distinct topic and added it to the topic dictionary.

\subsection{Feature Extraction}

\subsubsection{Acoustic Feature Extraction}

We extracted acoustic features using both traditional signal processing methods and deep learning techniques:

\paragraph{Traditional Acoustic Features:} Using OpenSMILE \cite{eyben2010opensmile}, we extracted the extended Geneva Minimalistic Acoustic Parameter Set (eGeMAPS) \cite{eyben2015geneva}, which consists of 88 low-level descriptors and their statistical functionals. These features capture various aspects of speech, including:
\begin{itemize}
    \item \textbf{Prosodic Features:} Parameters such as pitch (fundamental frequency) and energy, which reflect intonation, stress, and rhythm.
    \item \textbf{Spectral Features:} Characteristics like formant frequencies, spectral flux, and spectral slope, providing information about the timbre and quality of the voice.
    \item \textbf{Voice Quality Features:} Measures of jitter, shimmer, and harmonic-to-noise ratio, indicating variations in vocal fold vibration and breathiness.
\end{itemize}
We computed statistical functionals (e.g., mean, standard deviation, percentiles) over these low-level descriptors to summarize their distribution over time. This approach captures the temporal dynamics of speech which is crucial for identifying emotional cues and disordered speech patterns.

\paragraph{Deep Audio Representations:} We extracted deep acoustic embeddings through two different methods.
\begin{itemize}
    \item \textbf{DEEPSPECTRUM Features:} We converted audio signals into Mel-spectrograms, which represent the spectral content of the audio over time. These spectrograms were fed into a pre-trained DENSENET201 convolutional neural network \cite{huang2017densely}, as proposed in DEEPSPECTRUM \cite{amiriparian2017snore}. The network, trained on large image datasets, extracts high-level features that capture complex patterns in the spectrograms, effectively translating visual patterns into meaningful audio representations.
    \item \textbf{wav2vec 2.0 Embeddings:} We used wav2vec 2.0 \cite{baevski2020wav2vec}, a self-supervised model trained on raw audio data without labeled transcriptions. The model learns contextualized representations by predicting masked portions of the audio signal based on its context. By extracting 1,024-dimensional embeddings from the model's intermediate layers, we captured phonetic and prosodic information that reflects subtle variations in speech production.
\end{itemize}

\subsubsection{Text Feature Extraction}

We extracted textual features from transcribed speech using an in-house developed modular Python package called PELICAN (Preprocessing and Extraction of Linguistic Information for Computational Analysis). The various modules of PELICAN are designed to enable streamlined reproducible language processing. Using PELICAN's processing pipeline, we extracted 768-dimensional word embedding vectors with the multilingual language model XLM-RoBERTa-base \cite{conneau2019unsupervised}, leveraging its transformer-based architecture for contextual representation. These vectors helped us analyze semantic and syntactic structures at the granular level, which is essential for identifying language disruptions.


\subsection{Multimodal Fusion Techniques}

Integrating acoustic and linguistic features is central to our approach. To assess whether fusion provides additional predictive value beyond unimodal models, we compare our fusion strategies against text-only and audio-only models in the results section, we developed three fusion strategies: early fusion, late fusion, and Uncertainty-Aware Temporal Context Fusion (TCF). 

\textbf{Early Fusion} In the early fusion strategy, we combine acoustic and linguistic features at the utterance level to form a unified representation for each segment of speech. After aligning the audio and text modalities using forced alignment techniques, we concatenated the feature vectors from both modalities. This approach allows the model to learn joint representations that capture the interplay between linguistic content and acoustic delivery. By presenting both sets of features together, early fusion reveals  patterns that might remain hidden when analyzing each modality separately.

\textbf{Late Fusion} In the late fusion strategy, separate models process acoustic and linguistic features independently, each optimized for its respective modality. These models generate probability distributions over the target variable, which are then combined using dynamically weighted averaging.

Given predictions from the acoustic model \( p_A \) and the text model \( p_T \), the final prediction \( p_F \) is computed as:

\[
p_F = w_A \cdot p_A + w_T \cdot p_T, \quad \text{where} \quad w_A + w_T = 1.
\]

During cross-validation, the system dynamically adjusts the weights \( w_A \) and \( w_T \) in each fold based on validation loss, ensuring that the modality with lower error contributes more to the final prediction. After cross-validation, the final fusion weights are computed as the average across all folds, stabilizing their influence and preventing overfitting to a specific validation set.

\subsubsection{Temporal Context Fusion with Uncertainty Modeling} 

To integrate the temporal dynamics of both audio and text modalities effectively, we adopted an uncertainty-aware fusion strategy inspired by the Calibrated and Ordinal Latent Distributions (COLD) Fusion method proposed by Hazarika et al. \cite{hazarika2020misa, tellamekala2023cold}. Although the original COLD Fusion approach was designed for audiovisual data, we adapted it for our audio-text setting by modeling modality-specific uncertainties to improve fusion performance. By incorporating uncertainty modeling, we capture the temporal dynamics of both modalities while accounting for modality-wise aleatoric uncertainty.

\paragraph{Learning Unimodal Latent Distributions}

We quantified modality-wise uncertainty by learning unimodal latent distributions over the temporal context vectors of the audio and text modalities. Instead of representing the temporal context as deterministic embeddings, we modeled them as multivariate normal distributions:

\begin{align*}
h_i^M &\sim \mathcal{N}(\boldsymbol{\mu}_i^M, \boldsymbol{\Sigma}_i^M),
\end{align*}

where \(\boldsymbol{\mu}_i^M\) is the mean vector and \(\boldsymbol{\Sigma}_i^M\) is the covariance matrix at time step \(i\). We treat \(\boldsymbol{\Sigma}_i^M\) as a function of the input sample, which enables the model to learn heteroscedastic (input-dependent) aleatoric uncertainty \cite{kendall2017uncertainties}. This design is crucial for speech data, where noise levels vary with different speakers, settings, and utterances (e.g., background noise). If a particular utterance is noisier—due to muffled audio or rapid speech changes—the model can automatically assign a higher variance, reflecting lower confidence in that segment. While this primarily applies to the acoustic features, the text-based features are also affected by transcription quality. Errors in automatic speech recognition, missing words, or misinterpretations can introduce inconsistencies in the textual representation, leading to potential loss of linguistic information.

\paragraph{Cross-Modal Attention} To further refine the interaction between the two modalities, we applied cross-modal attention. This mechanism allows the model to dynamically focus on relevant parts of one modality based on the context provided by the other. For example, it treats audio hidden states as queries and text hidden states as keys and values. The attention mechanism calculates how much each audio unit should focus on the text and vice versa, using a weighted sum to form context-aware representations for both modalities. This process enables the model to align important features between audio and text, making the fusion more robust to timing discrepancies.

\paragraph{Deriving Fusion Weights}

To determine each modality's contribution to the fused representation, we calculated fusion weights based on the inverse of the variance norms of the unimodal latent distributions. The idea is that a modality with lower uncertainty (smaller variance norm) should have a greater influence on the fused output \cite{tel}. The fusion weights for a modality ($w_i^M$) at time step $i$ were calculated as:

\begin{align*}
    w_i^A &= \frac{\| \boldsymbol{\Sigma}_i^T \|_2}{\| \boldsymbol{\Sigma}_i^A \|_2 + \| \boldsymbol{\Sigma}_i^T \|_2}, \\
    w_i^T &= \frac{\| \boldsymbol{\Sigma}_i^A \|_2}{\| \boldsymbol{\Sigma}_i^A \|_2 + \| \boldsymbol{\Sigma}_i^T \|_2}.
\end{align*}

This assigns a higher weight to the modality with lower variance (indicating lower uncertainty) in the fusion process.

The fused context vector $h_i^{AT}$ is then computed as a weighted sum of the unimodal context mean vectors:

\begin{equation*} 
h_i^{AT} = w_i^A \cdot \mu_i^A + w_i^T \cdot \mu_i^T, 
\end{equation*}

\paragraph{Constraints}

To ensure that the variance vectors accurately represent modality uncertainties, we used two key constraints during training: Calibration and Ordinality. These constraints guide the model to align the variances with the prediction errors of each modality, effectively modeling uncertainty and regulating each modality's contribution to the fused representation \cite{tellamekala2023cold}.

\paragraph{Calibration Constraint}

The calibration constraint encourages a correlation between the variance norms and the prediction errors of each modality. This ensures that the variance norms reflect the expected estimation errors, making the model’s uncertainty measure reliable. For each modality, we calculate the prediction errors \( e_i^A \) and \( e_i^T \) as follows:

\begin{equation}
    e_i^A = | y_i^A - y_i^\ast |^2, \quad e_i^T = | y_i^T - y_i^\ast |^2,
\end{equation}

where \( y_i^A \) and \( y_i^T \) are the unimodal predictions for audio and text, respectively, and \( y_i^\ast \) is the ground truth label at time step \( i \). By aligning the variance norms \( \| \boldsymbol{\Sigma}_i^A \|_2 \) and \( \| \boldsymbol{\Sigma}_i^T \|_2 \) with the prediction errors, we ensure that higher variance values correspond to higher prediction errors, thereby capturing the uncertainty associated with each modality.

\paragraph{Ordinality Constraint}

The ordinality constraint enforces a ranking among uncertainties across modalities. This ranking ensures that the modality with higher uncertainty (larger variance norm) contributes less to the fusion process. We do this by aligning the ranks of the variance norms with those of the prediction errors. Formally, this constraint is:

\begin{equation}
    \text{If } e_i^A > e_i^T, \text{ then } \| \boldsymbol{\Sigma}_i^A \|_2 > \| \boldsymbol{\Sigma}_i^T \|_2.
\end{equation}

This constraint ensures that when the prediction error of the audio modality exceeds that of the text modality, the audio variance norm is correspondingly larger, indicating lower reliability.

\paragraph{Implementation of Constraints}

To implement these constraints, we minimize the Kullback-Leibler (KL) divergence between the softmax distributions of the prediction errors and the variance norms. Specifically, we define the combined Loss \( L_{\text{CO}} \) as:

\begin{equation}
    L_{\text{CO}} = \text{KL}( P_e \parallel P_\sigma ) + \text{KL}( P_\sigma \parallel P_e ),
\end{equation}

where \( P_e \) and \( P_\sigma \) are the softmax-normalized distributions of the prediction errors and variance norms, respectively:

\begin{equation}
    P_e = \text{softmax}( -\mathbf{e} ), \quad P_\sigma = \text{softmax}( -\boldsymbol{\Sigma}^2 ).
\end{equation}

The negative sign guarantees that higher errors or variances correspond to lower probabilities. By minimizing \( L_{\text{CO}} \), we encourage the variance norms to be proportional to the prediction errors, enabling the model to reliably represent uncertainty across both modalities.


\subsection{Tasks and Evaluation}

We designed our study to include both classification and regression tasks, with the goal of capturing the full range of schizotypal traits using speech and language analysis.

\textbf{Classification Tasks} We grouped participants into three categories: individuals with low schizotypy, those with high schizotypy, and patients diagnosed with schizophrenia. The low and high schizotypy groups included healthy individuals, allowing us to examine how varying levels of schizotypal traits present in speech patterns even without a clinical diagnosis. We analyzed each session type (Semi-structured Autobiographical Interview, Thematic Apperception Test (TAT), The PANSS Clinical Interview and DISCOURSE session) separately to understand how different interaction contexts influenced the models' ability to distinguish among these groups.

\textbf{Regression Tasks} We predicted clinical scores from the MSS, the O-LIFE, and the PANSS. By linking speech and language features to the severity of clinical symptoms, we quantified the relationship between vocal and linguistic patterns and schizotypal traits. We used 5-Fold Cross-Validation to evaluate the robustness and generalizability of our models, ensuring that training and testing were conducted across different participants and session types.

We also compared model performance across different interaction types. This comparison allowed us to determine how the setting influences predictive accuracy and to identify which contexts provide the most reliable insights into clinical symptoms and levels of schizotypy.

\textbf{Evaluation Metrics}
We assessed model performance using accuracy and F1-score for classification tasks, and root mean squared error (RMSE) for regression tasks. Additionally, we computed expected calibration error (ECE) to evaluate the reliability of predicted probabilities in classification models. We used ECE to determine how well predicted probabilities align with observed frequencies in classification tasks (evaluate whether our uncertainty estimates match true model errors). These metrics offered a detailed view of how well the models predicted group classifications and assessment scores. 

For baseline comparisons, we used Random Forest (RF), Linear Discriminant Analysis (LDA), and Support Vector Machines (SVM). Random Forest, with its ensemble of decision trees, captures diverse decision boundaries and handles high-dimensional data effectively. Linear Discriminant Analysis finds a linear combination of features that best separates the classes. Support Vector Machines find the optimal hyperplane that maximizes the margin between classes, accommodating both linear and nonlinear relationships through kernel functions. These baseline models provided a reference point for evaluating the performance of more advanced methods, such as wav2vec and RoBERTa for speech and language tasks, and the fusion models for multimodal integration.

\subsection{Interpretation of Model Predictions}

To understand our model's predictions, we use SHAP (SHapley Additive exPlanations) values, which quantify each input feature's contribution to the output. Calculating exact Shapley values is computationally intensive for deep models, so we apply Deep SHAP, an efficient approximation based on DeepLIFT \cite{shrikumar2017learning}.

Deep SHAP estimates each feature's contribution by comparing neuron activations to those from a baseline input. The contribution \(C_{ij}\) of feature \(i\) to output neuron \(j\) is calculated as:

\[
C_{ij} = (x_i - x_i') \cdot \frac{\partial y_j}{\partial x_i}
\]

Here, \(x_i\) is the actual input value, \(x_i'\) is the baseline value, and \(\frac{\partial y_j}{\partial x_i}\) is the gradient of the output with respect to \(x_i\). We approximate the SHAP value \(\phi_i\) for feature \(i\) by summing these contributions:

\[
\phi_i \approx \sum_j C_{ij}
\]

To clarify significance in SHAP values, we establish confidence intervals using \( k \)-fold cross-validation. We train the model across \( k \) folds and compute SHAP values on validation sets. The 2.5th and 97.5th percentiles of these values define the 95\% confidence interval (CI) for each feature. A feature is considered significant if its CI excludes zero, indicating a stable contribution to predictions across folds.

Some features with low SHAP values may still be significant if their contributions remain consistent, whereas features with high SHAP values but large variance may not reach significance if their effect fluctuates, causing the CI to overlap with zero.

To improve comparability across interaction types, we standardize SHAP values by normalizing each feature’s absolute SHAP value relative to the total SHAP values within each model. We then compute average normalized SHAP values across datasets. If datasets vary in size, we weight the SHAP values by the number of instances \( N_d \) in each dataset. This approach ensures that identified features reflect stable and meaningful contributions across different contexts.

To interpret the word embeddings in our language model, we replace each word with its corresponding category from the Linguistic Inquiry and Word Count (LIWC) lexicon similar to \cite{kilic2022incorporating}. When a word belongs to multiple categories, we combine them to capture the full range of its meaning. Words absent from the LIWC lexicon receive a "None" label. This method shifts the model's focus to linguistic features tied to emotional and cognitive processes, enhancing how we understand the underlying patterns in the embeddings.

\begin{table*}[t]
\centering
\rowcolors{2}{LightGrey}{white}  
\begin{tabularx}{\textwidth}{lX|c>{\columncolor{LightBlue}}c>{\columncolor{LightBlue}}c|c>{\columncolor{LightBlue}}c>{\columncolor{LightBlue}}c|c>{\columncolor{LightBlue}}c>{\columncolor{LightBlue}}c|c>{\columncolor{LightBlue}}c>{\columncolor{LightBlue}}c}
\toprule
& \textbf{Interaction Type} & \multicolumn{3}{c}{ \textbf{Interview}} & \multicolumn{3}{c}{\textbf{TAT}} & \multicolumn{3}{c}{\textbf{PANSS}} & \multicolumn{3}{c}{\textbf{DISCOURSE}} \\
& \textbf{Metric} & \textbf{Acc.↑} & \textbf{F1↑} & \textbf{ECE↓} & \textbf{Acc.↑} & \textbf{F1↑} & \textbf{ECE↓} & \textbf{Acc.↑} & \textbf{F1↑} & \textbf{ECE↓} & \textbf{Acc.↑} & \textbf{F1↑} & \textbf{ECE↓} \\
\midrule
\textbf{Speech} & Random Forest & 74 & 74 & 8.0e-2 & 60 & 60 & 9.0e-2 & 69 & 68 & 8.0e-2 & 55 & 54 & 8.0e-2 \\
& LDA & 59 & 58 & 14.0e-2 & 53 & 53 & 15.1e-2 & 60 & 60 & 16.2e-2 & 51 & 50 & 10.0e-2 \\
& SVM & 68 & 68 & 11.0e-2 & 57 & 56 & 14.0e-2 & 68 & 67 & 15.0e-2 & 55 & 55 & 8.0e-2 \\
& GRU & \textbf{77} & \textbf{76} & 11.0e-2 & 63 & 63 & 11.1e-2 & 71 & 71 & 14.2e-2 & 62 & \textbf{63} & 9.0e-2 \\
& DEEPSPECTRUM & 76 & 76 & 8.0e-2 & 64 & 64 & \textbf{8.0e-2} & 73 & 73 & 9.0e-2 & 59 & 58 & 7.0e-2 \\
& wav2vec-Large & \textbf{77} & \textbf{76} & \textbf{7.0e-2} & \textbf{64} & \textbf{64} & 9.0e-2 & \textbf{73} & \textbf{73} & \textbf{9.0e-2} & \textbf{60} & 60 & \textbf{6.0e-2} \\
\midrule
\textbf{Language} & Random Forest & 77 & 76 & 9.0e-2 & 63 & 63 & 8.0e-2 & 73 & 73 & 7.0e-2 & 67 & 67 & 7.0e-2 \\
& LDA & 64 & 63 & 11.0e-2 & 57 & 57 & 11.1e-2 & 62 & 63 & 14.2e-2 & 63 & 63 & 12.0e-2 \\
& SVM & 68 & 68 & 8.0e-2 & 60 & 60 & 9.0e-2 & 66 & 65 & 10.0e-2 & 63 & 63 & 15.0e-2 \\
& GRU & 80 & 79 & 10.0e-2 & 70 & 69 & 10.1e-2 & 77 & 77 & 12.2e-2 & 68 & 68 & 9.0e-2 \\
& Fine-tuned BERT & \textbf{81} & \textbf{80} & \textbf{7.0e-2} & 71 & 70 & \textbf{7.0e-2} & 78 & 77 & 8.0e-2 & 68 & 68 & 7.0e-2 \\
& Fine-tuned RoBERTa & \textbf{81} & \textbf{80} & \textbf{5.0e-2} & \textbf{73} & \textbf{73} & 8.0e-2 & \textbf{78} & \textbf{78} & \textbf{7.0e-2} & \textbf{68} & 67 & 8.0e-2 \\
\bottomrule
\end{tabularx}
\caption{Performance of Speech and Language Models across different interaction types of Semi-structured Autobiographical Interview (Interview), Thematic Apperception Test (TAT), The PANSS Clinical Interview (PANSS) and DISCOURSE session in the classification task. Metrics: Accuracy (Acc.↑), F1 Score (F1↑), and Expected Calibration Error (ECE↓).}
\end{table*}

\section{Results}
\subsection{Performance of Speech and Language Models in Classification}
In Table 1, we study how different models perform across various interaction types using both speech and text data. These results show classification tasks intended to distinguish individuals into three classes based on symptom severity: low schizotypy, high schizotypy, and patients with psychosis. We compared the models by examining accuracy and ECE. For speech-based tasks, the wav2vec-Large model consistently outperforms others, reaching the highest accuracy (77\%) in Semi-structured and PANSS sessions (73\%). The lower ECE scores of wav2vec-Large (7.0e-2 for Semi-structured and 9.0e-2 for PANSS) show a more reliable calibration in predicting clinical outcomes from acoustic data. 
The model’s ability to capture subtle vocal cues aids in accurately classifying symptom severity, supporting the link between acoustic features and underlying psychiatric symptoms.

DEEPSPECTRUM follows closely, although its slightly higher error rate signals potential limitations in its ability to generalize across varying session dynamics. 

In comparison, traditional models like Random Forest and SVM fall behind, particularly in less structured settings such as DISCOURSE, where accuracy drops below 60\%. The reliance on decision boundaries that fail to capture the full variability in speech may account for this. GRU models show some improvement over baselines but still lag behind wav2vec.

For text-based models, the fine-tuned RoBERTa outperforms all others, achieving 81\% accuracy in Semi-structured settings and maintaining a strong performance across TAT (73\%) and PANSS sessions (78\%). With an ECE of 5.0e-2 in Semi-structured interactions, RoBERTa shows both accuracy and confidence in classifying different groups. This reflects the strength of transformer models in understanding the linguistic disruptions characteristic of schizophrenia. BERT closely follows, and its slightly higher ECE in TAT (7.0e-2) suggests that subtle shifts in word choice, perhaps more pronounced in spontaneous speech, are critical for identifying symptoms.

LDA and SVM, though simpler, perform adequately in structured interactions like PANSS but struggle in more open-ended sessions, where DISCOURSE patterns become less predictable. Their limitations show the importance of contextual and sequential understanding in clinical language tasks, where traditional linear classifiers fail to capture the depth of disorganized thought patterns that advanced models handle with more precision.

\subsection{Performance of Multimodal Fusion Models}

Table 2 shows the performance of our multimodal fusion models across different interaction types. The TCF model outperforms all others in both accuracy and calibration error, reaching 83\% accuracy in Semi-structured and TAT sessions, and maintaining the lowest ECE values across the board. This reflects TCF’s ability to capture the temporal and modality-specific uncertainty, a key aspect of our hypothesis that speech disruptions and schizotypal traits can be better predicted by uncertainty-aware models.

Compared to simpler fusion strategies like Early and Late Fusion, TCF integrates both acoustic and linguistic information more effectively, especially in structured sessions like PANSS, where it achieves an accuracy of 77\% with an ECE of 3.8e-2. These results validate our approach of incorporating modality-wise uncertainty into the fusion process. It shows that uncertainty-aware models capture the variability and the subtle fluctuations in speech and language across the schizophrenia spectrum, which traditional fusion methods fail to handle.

Late Fusion, though more stable than Early Fusion, underperforms in less structured settings like DISCOURSE, where accuracy drops to 63\%. This highlights the need for real-time integration of multimodal data, especially when speech is less predictable, as is often the case in spontaneous interactions. Early Fusion, despite its simplicity, struggles the most, particularly in DISCOURSE sessions where accuracy plummets to 59\%. The lack of temporal awareness in Early Fusion clearly limits its capacity to handle the dynamic nature of speech patterns critical for detecting schizotypal traits.

The superiority of TCF directly supports our hypothesis that advanced, uncertainty-aware models are essential for decoding the complex speech and linguistic patterns inherent in schizophrenia and schizotypy. By modeling uncertainty, TCF not only improves predictive accuracy but also aligns with our objective of monitoring symptom variability across clinical and non-clinical populations.

Table 3 highlights the performance of TCF models across different interaction types in the transfer learning setting. When trained on Semi-structured data, the model performs strongly in both TAT and DISCOURSE, achieving 73\% and 60\% accuracy, respectively. The lower ECE values in these settings reflect the model’s ability to generalize beyond the controlled Semi-structured interviews, where specific prompts allow for consistent speech markers. This shows the hypothesis that our approach captures variability, even when tested in less predictable settings.

\begin{table*}[t]
\centering
\rowcolors{2}{LightGrey}{white}
\begin{tabularx}{\textwidth}{lX|ccc|ccc|ccc|ccc}
\toprule
\textbf{Models} & \textbf{Interaction Type} & \multicolumn{3}{c}{\textbf{Interview}} & \multicolumn{3}{c}{\textbf{TAT}} & \multicolumn{3}{c}{\textbf{PANSS}} & \multicolumn{3}{c}{\textbf{DISCOURSE}} \\
                & \textbf{Metric} & \textbf{Acc.↑} & \textbf{F1↑} & \textbf{ECE↓} & \textbf{Acc.↑} & \textbf{F1↑} & \textbf{ECE↓} & \textbf{Acc.↑} & \textbf{F1↑} & \textbf{ECE↓} & \textbf{Acc.↑} & \textbf{F1↑} & \textbf{ECE↓} \\
\midrule
Early Fusion    & & 0.74 & 0.73 & 7.0e-2 & 0.73 & 0.72 & 9.0e-2 & 0.73 & 0.73 & 9.5e-2 & 0.59 & 0.58 & 10.0e-2 \\
Late Fusion     & & 0.80 & 0.79 & 7.0e-2 & 0.78 & 0.77 & 7.0e-2 & 0.74 & 0.73 & 8.5e-2 & 0.63 & 0.62 & 9.0e-2 \\
Context Fusion  & & 0.81 & 0.80 & 6.5e-2 & 0.80 & 0.79 & 7.0e-2 & 0.76 & 0.75 & 7.5e-2 & 0.66 & 0.65 & 8.0e-2 \\
TCF      & & \textbf{0.83} & \textbf{0.83} & \textbf{4.5e-2} & \textbf{0.83} & \textbf{0.82} & \textbf{5.0e-2} & \textbf{0.77} & \textbf{0.77} & \textbf{3.8e-2} & \textbf{0.68} & \textbf{0.67} & \textbf{6.0e-2} \\
\bottomrule
\end{tabularx}
\caption{Performance of Multimodal Fusion Models: Early, Late, Context, and Temporal Context Fusion across interaction types. Metrics: Accuracy (Acc.↑), F1 Score (F1↑), and Expected Calibration Error (ECE↓).}
\end{table*}

Training on Photo-elicitation (TAT), a more interview-style interaction with task-based stimulus, shows weaker generalization. While TAT itself achieves moderate accuracy, transferring to PANSS or DISCOURSE settings shows a noticeable drop in performance, with 46\% accuracy in PANSS. This suggests that the elicitation-based structure may not prepare the model for handling the broader, symptom-focused exchanges in clinical tests like PANSS, where psychotic symptoms are explicitly targeted and require a different kind of speech comprehension.

Models trained on PANSS data, however, show a stronger transfer to Semi-structured and DISCOURSE settings, achieving 71\% accuracy in Semi-structured sessions and 60\% in DISCOURSE. PANSS, as the clinical gold standard for psychosis, captures core symptoms more rigorously, which helps the model adapt to other structured and task-driven interactions. This supports the idea that more symptom-focused sessions provide richer data for predicting schizotypal speech patterns across different contexts.

Training on DISCOURSE, despite its structured task-based design, does not generalize well to clinical tests like PANSS, where accuracy drops to 55\%. This suggests that while DISCOURSE is useful for capturing certain cognitive patterns, it lacks the depth of symptom-focused interactions, leading to weaker performance in more symptom-driven sessions like PANSS. These results reinforce our hypothesis that the context of interaction plays a critical role in determining the success of speech-based models in schizophrenia research.

\begin{figure*}[htp]
    \centering
    \includegraphics[width=0.48\textwidth, height=0.20\textheight]{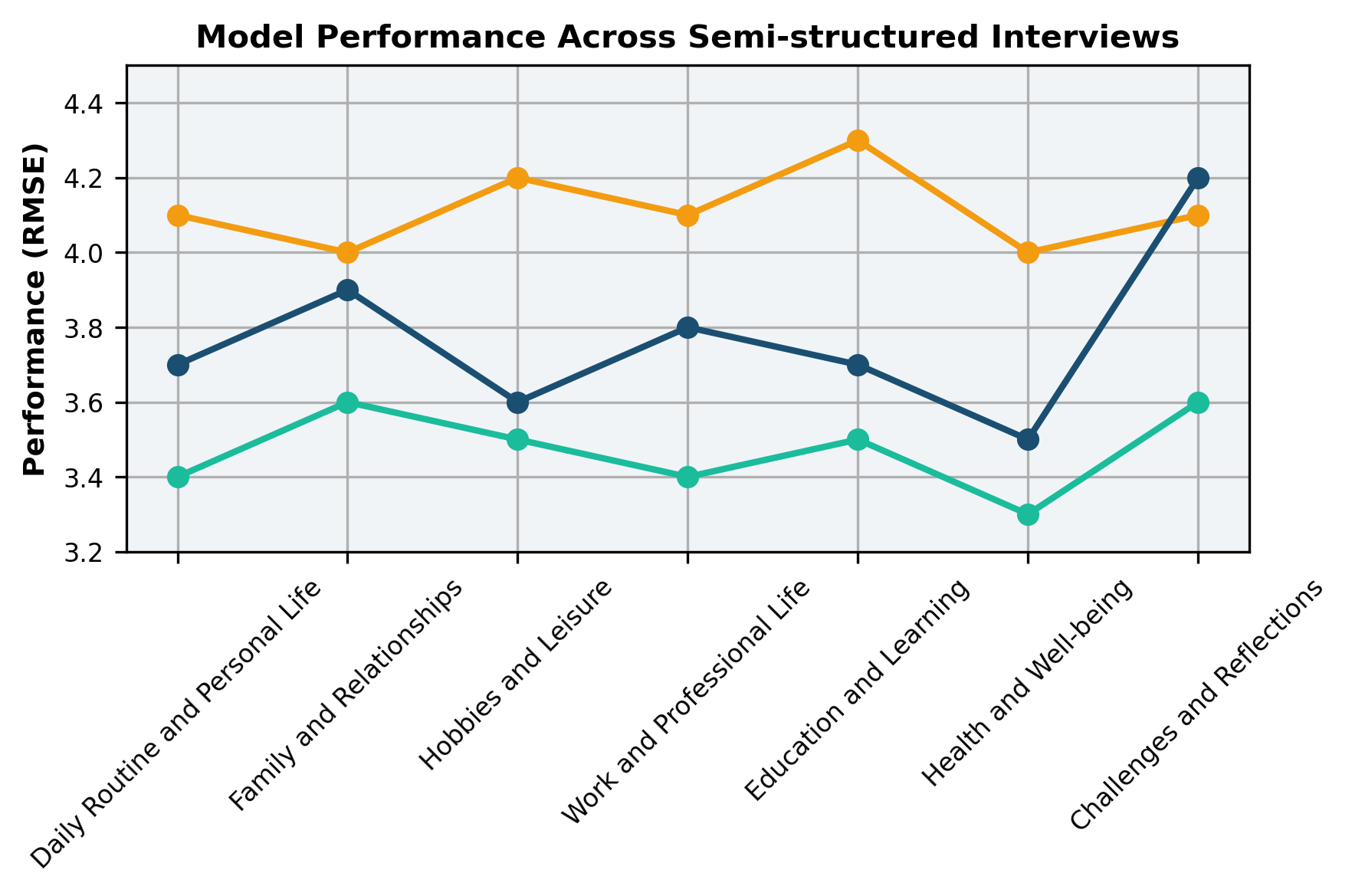}
    \hfill
    \includegraphics[width=0.48\textwidth, height=0.20\textheight]{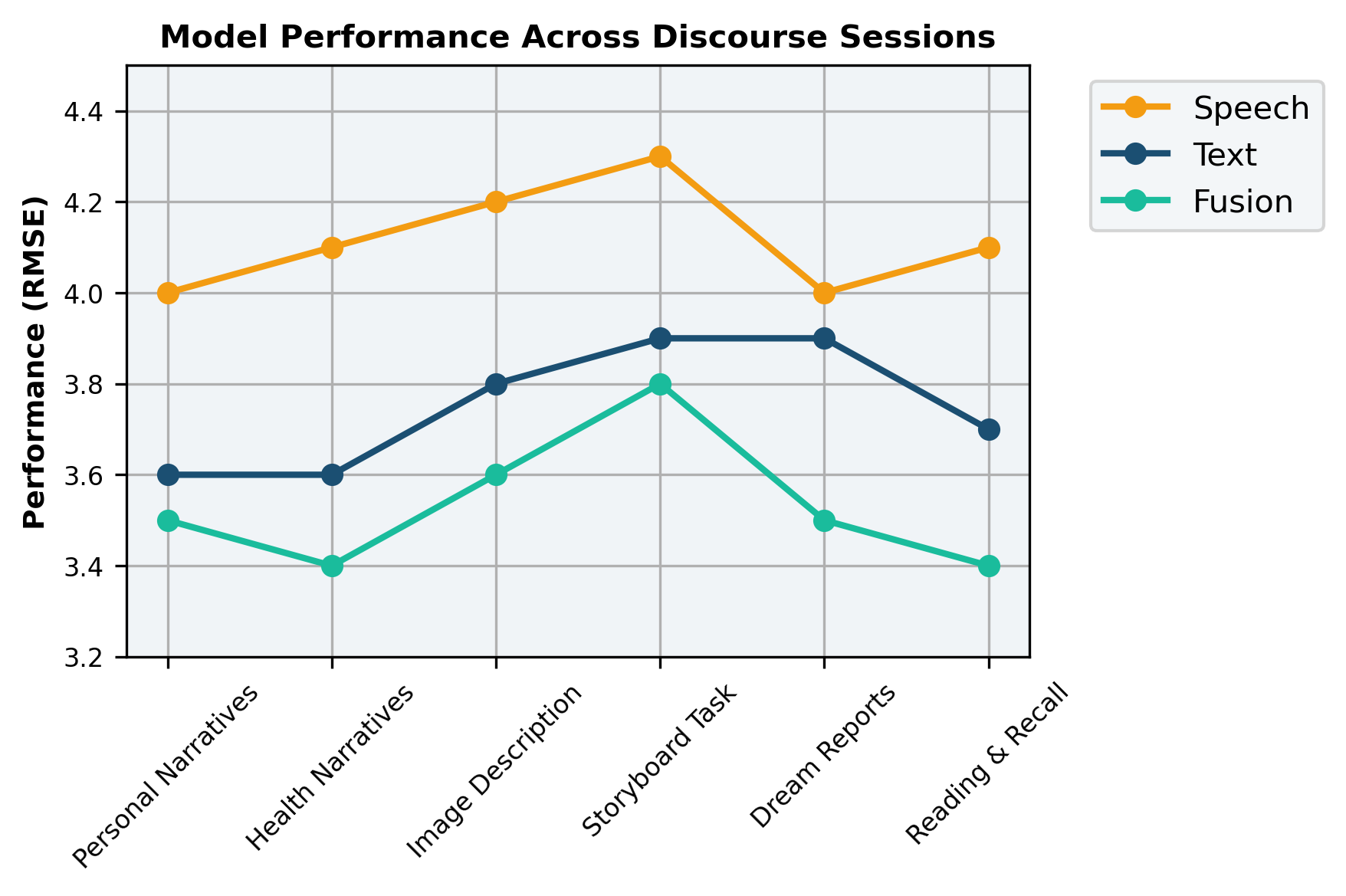}

    \vspace{-0.8em} 

    \includegraphics[width=0.48\textwidth, height=0.20\textheight]{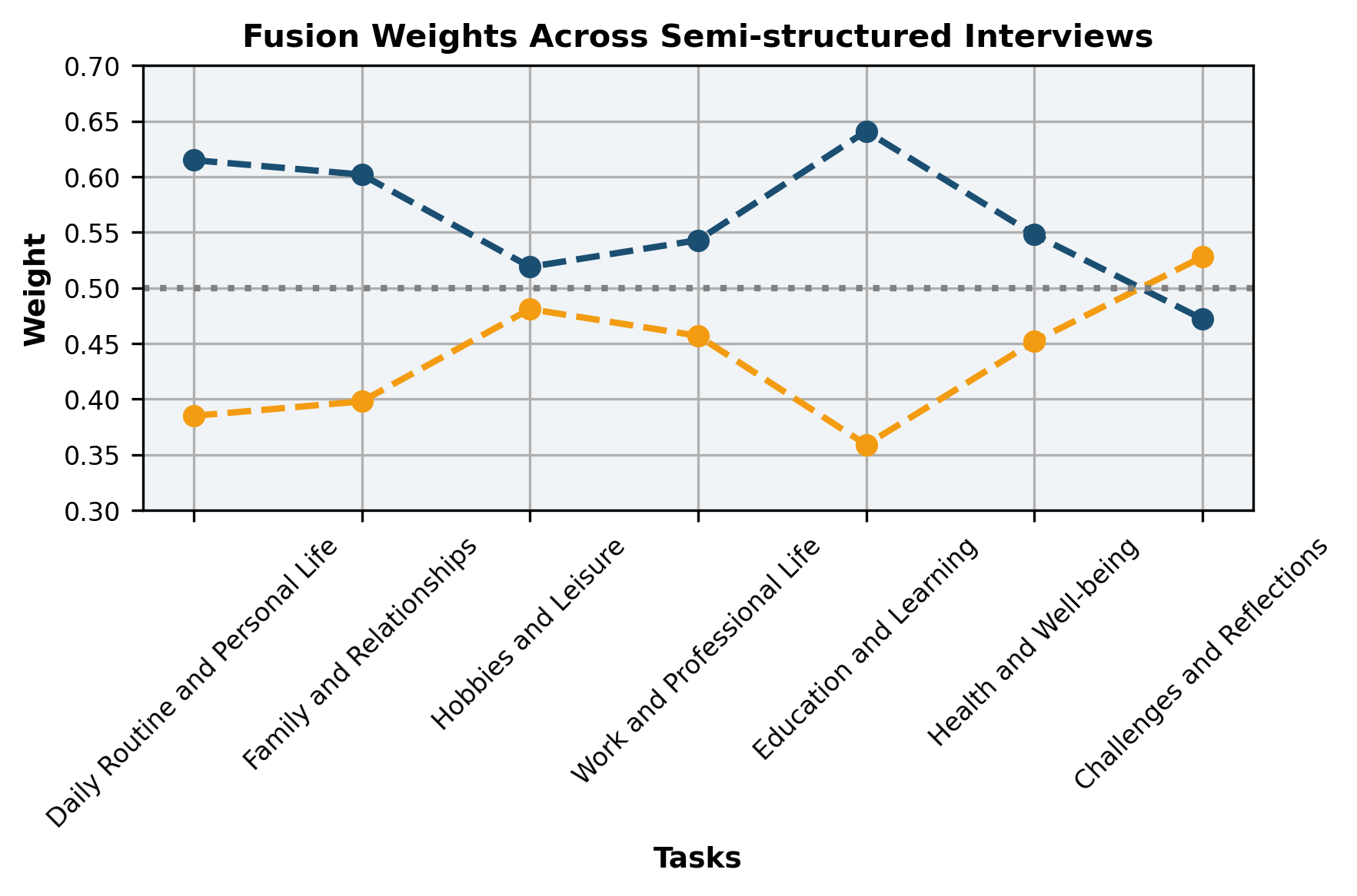}
    \hfill
    \includegraphics[width=0.48\textwidth, height=0.20\textheight]{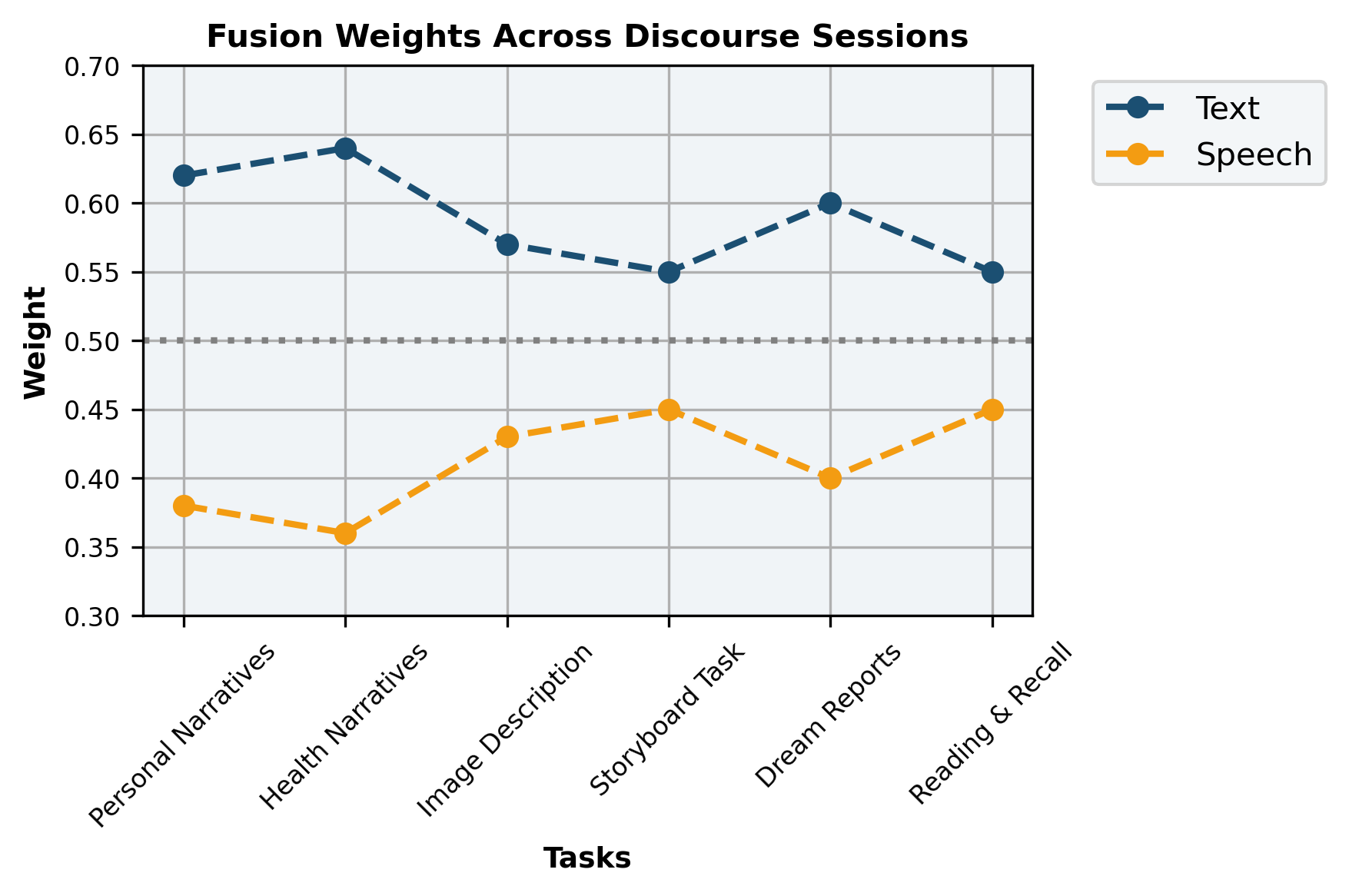}

    \caption{Comparison of model performance and fusion weights across different tasks and modalities. The top row shows the RMSE values for speech, text, and fusion models across semi-structured interviews and discourse sessions. The bottom row shows the fusion weights of speech and text modalities, highlighting the balance between modalities.}
    \label{fig:model_performance_weights}
\end{figure*}

\begin{table*}[t]
\centering
\rowcolors{2}{LightGrey}{white}
\begin{tabularx}{\textwidth}{lX|ccc|ccc|ccc|ccc}
\toprule
\textbf{TCF Models} & & \multicolumn{3}{c}{\textbf{Test: Interview}} & \multicolumn{3}{c}{\textbf{Test: TAT}} & \multicolumn{3}{c}{\textbf{Test: PANSS}} & \multicolumn{3}{c}{\textbf{Test: DISCOURSE}} \\
                            & & \textbf{Acc.↑} & \textbf{F1↑} & \textbf{ECE↓} & \textbf{Acc.↑} & \textbf{F1↑} & \textbf{ECE↓} & \textbf{Acc.↑} & \textbf{F1↑} & \textbf{ECE↓} & \textbf{Acc.↑} & \textbf{F1↑} & \textbf{ECE↓} \\
\midrule
\textbf{Train: Interview} & & -- & -- & -- & \textbf{0.73} & \textbf{0.73} & \textbf{9.5e-2} & 0.44 & 0.44 & 18.5e-2 & \textbf{0.60} & \textbf{0.60} & \textbf{14.5e-2} \\
\textbf{Train: TAT} & & 0.67 & 0.67 & 12.0e-2 & -- & -- & -- & 0.46 & 0.45 & 19.5e-2 & 0.54 & 0.54 & 16.0e-2 \\
\textbf{Train: PANSS} & & \textbf{0.71} & \textbf{0.70} & \textbf{10.0e-2} & 0.69 & 0.68 & 12.5e-2 & -- & -- & -- & 0.60 & 0.59 & 17.0e-2 \\
\textbf{Train: DISCOURSE} & & 0.65 & 0.64 & 15.0e-2 & 0.62 & 0.61 & 16.8e-2 & \textbf{0.55} & \textbf{0.54} & \textbf{17.5e-2} & -- & -- & -- \\
\bottomrule
\end{tabularx}
\caption{Cross-Corpus Performance of TCF Fusion Models for different interaction types. Metrics: Accuracy (Acc.↑), F1 Score (F1↑), and Expected Calibration Error (ECE↓) for training and testing across different interaction types.}
\end{table*}

\begin{figure*}[htp]
    \centering

    \includegraphics[width=0.48\textwidth]{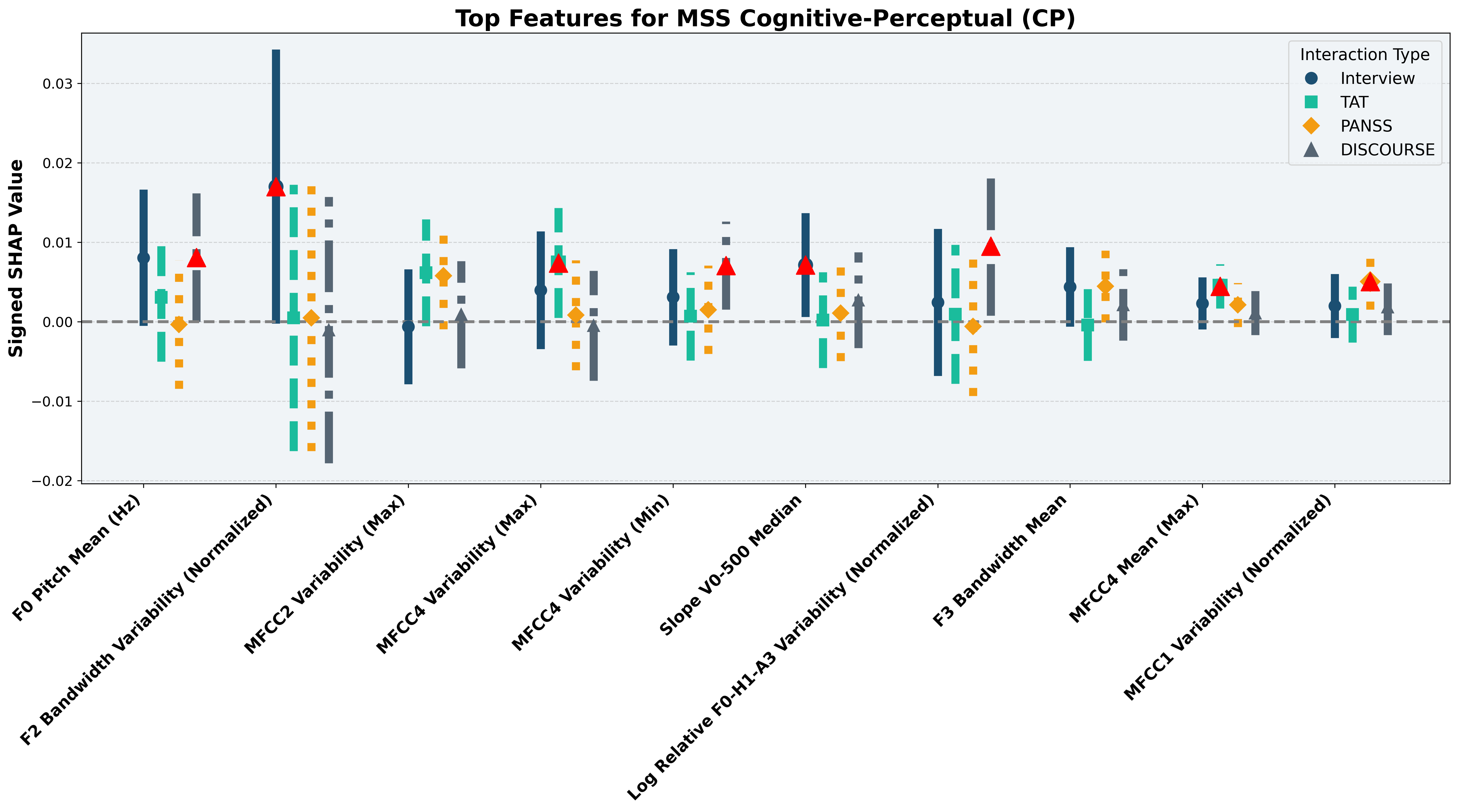}
    \hfill
    \includegraphics[width=0.48\textwidth]{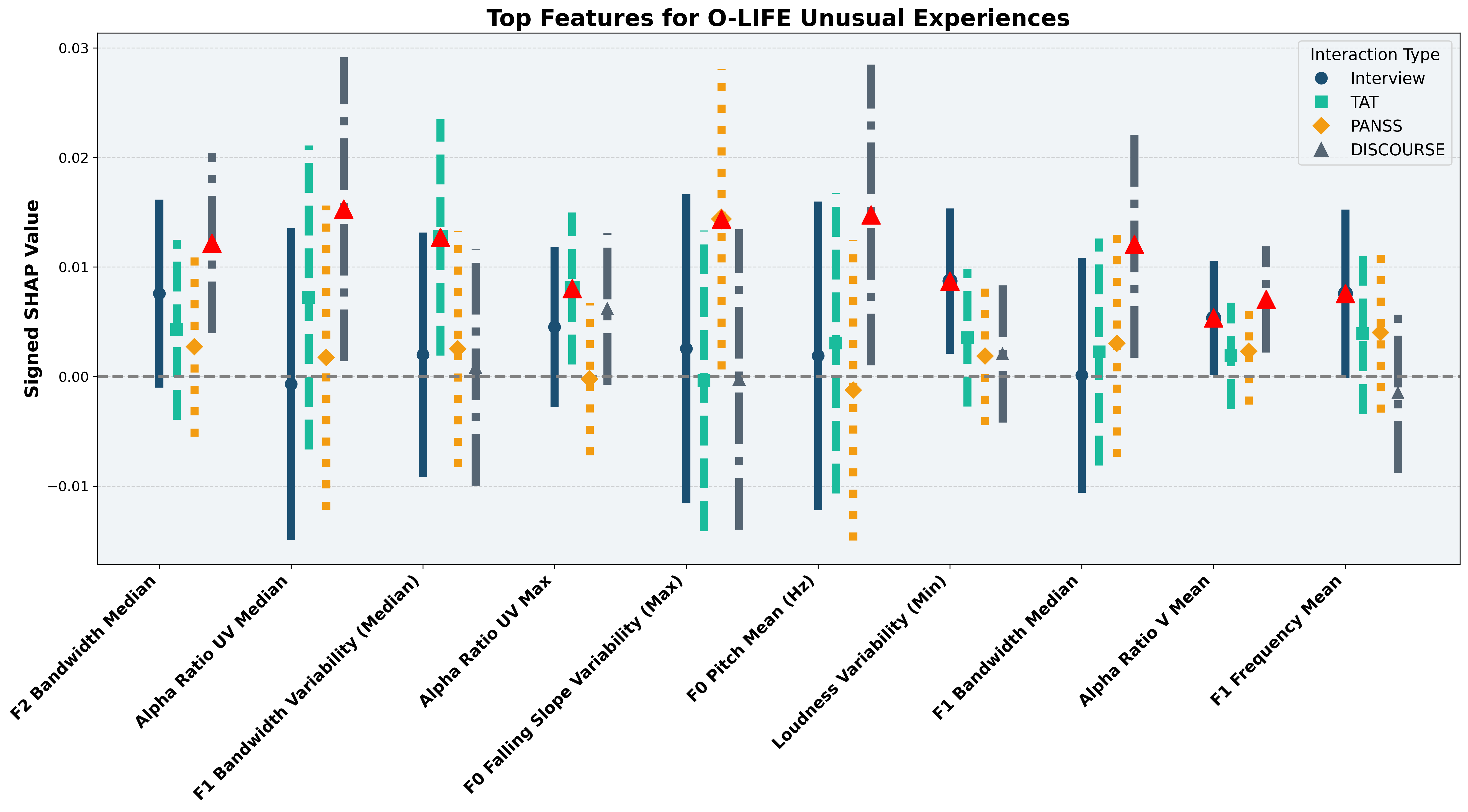}

    \vspace{0.5em} 
    \includegraphics[width=0.48\textwidth]{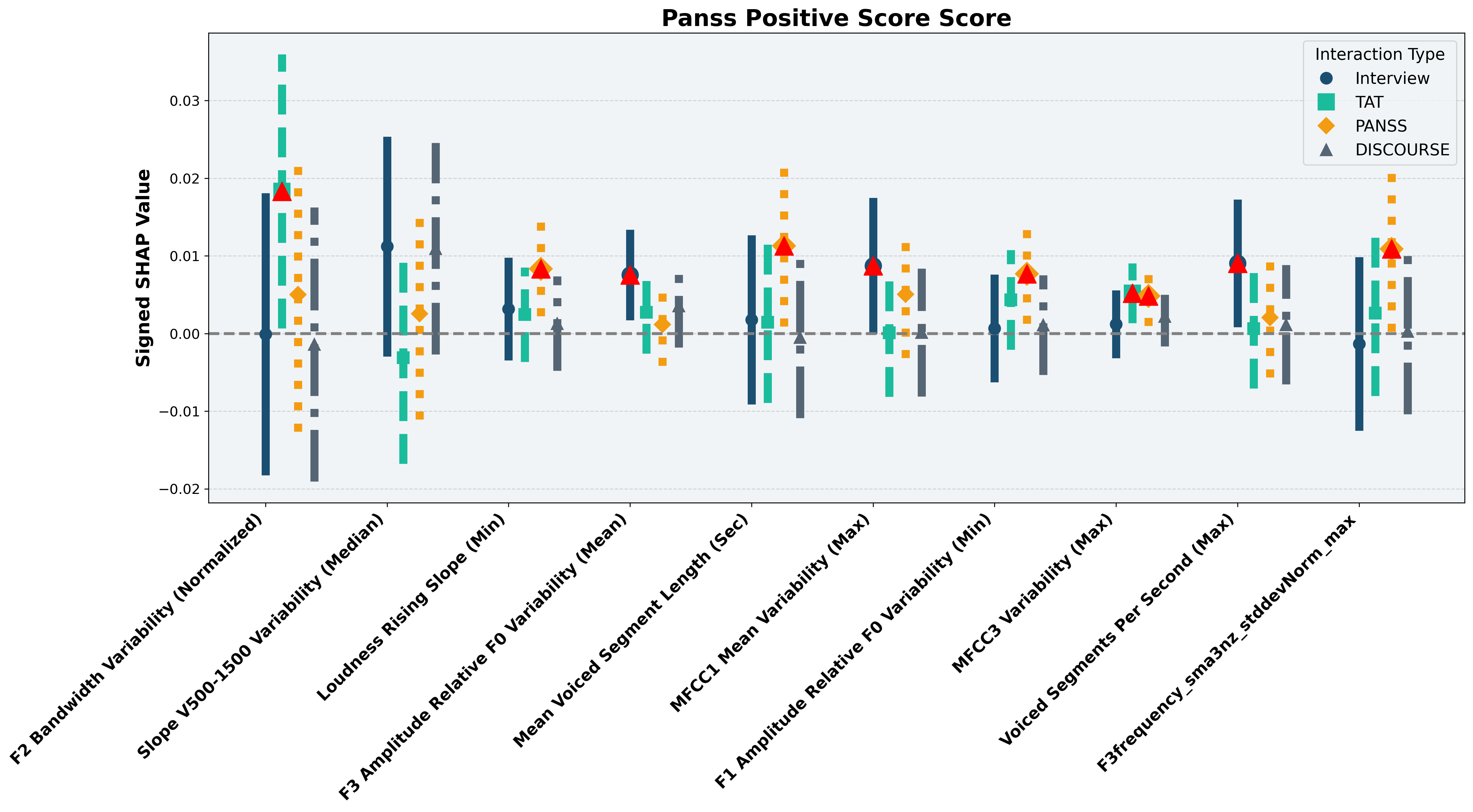}
    \hfill
    \includegraphics[width=0.48\textwidth]{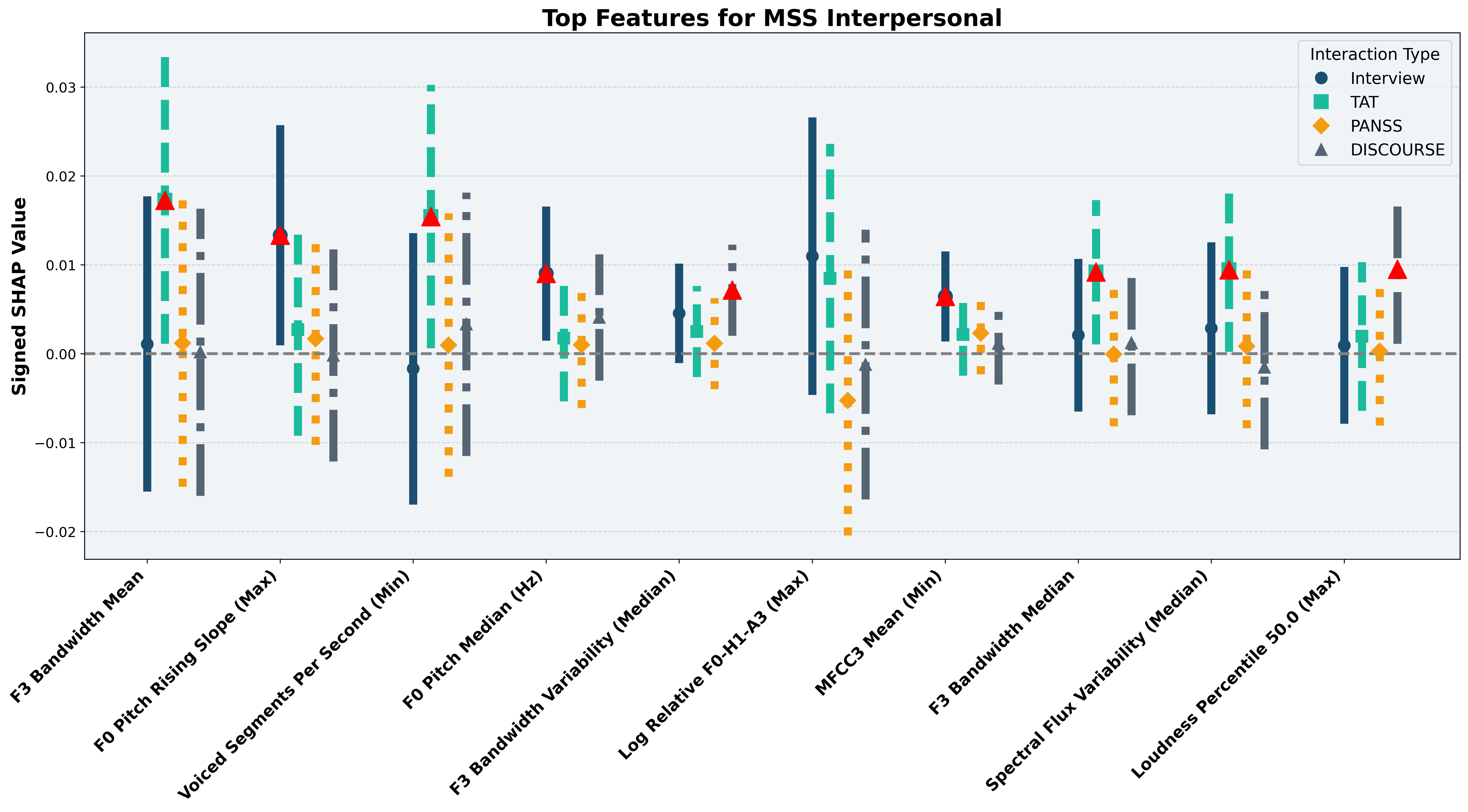}

    \vspace{0.5em}
    \includegraphics[width=0.48\textwidth]{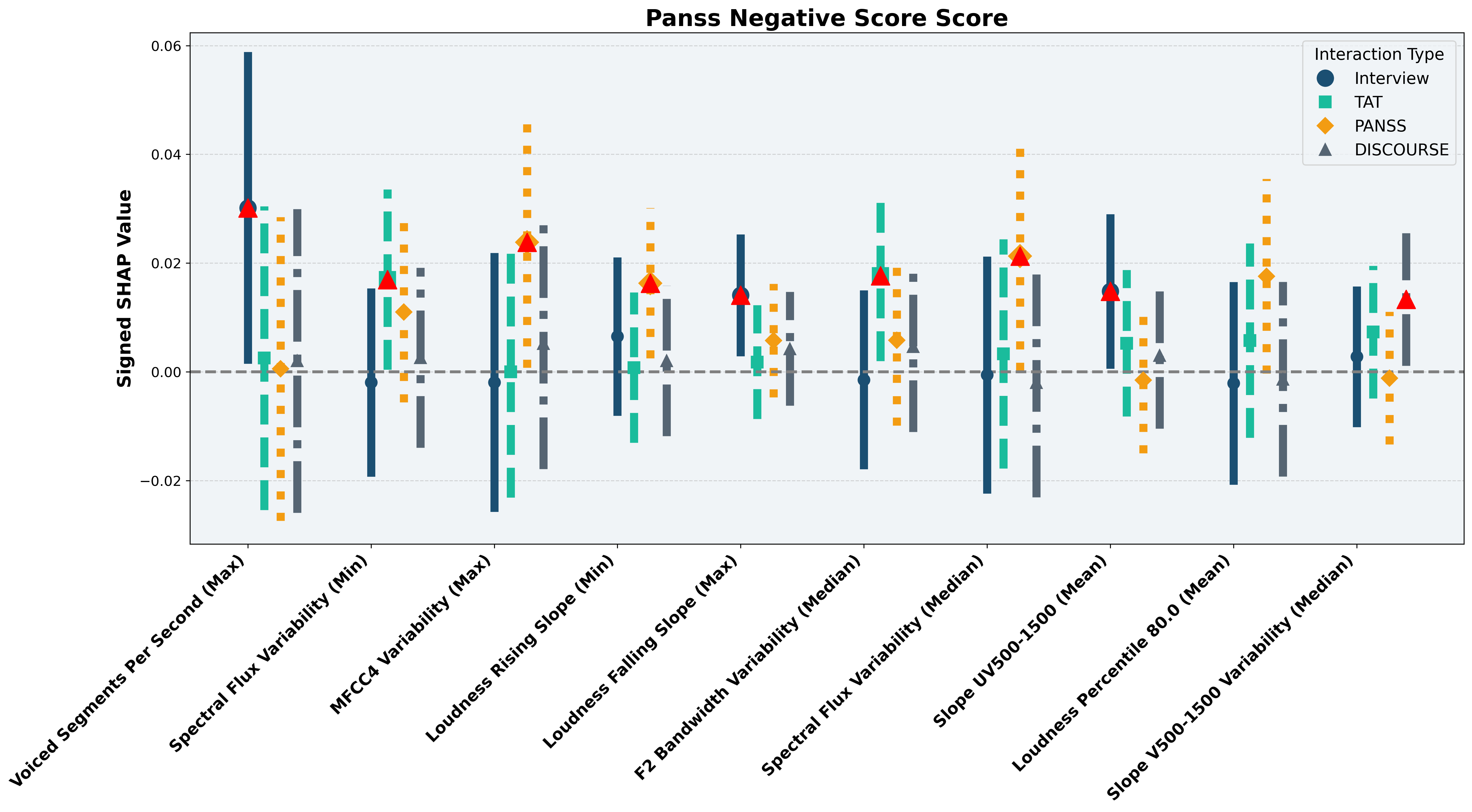}
    \hfill
    \includegraphics[width=0.48\textwidth]{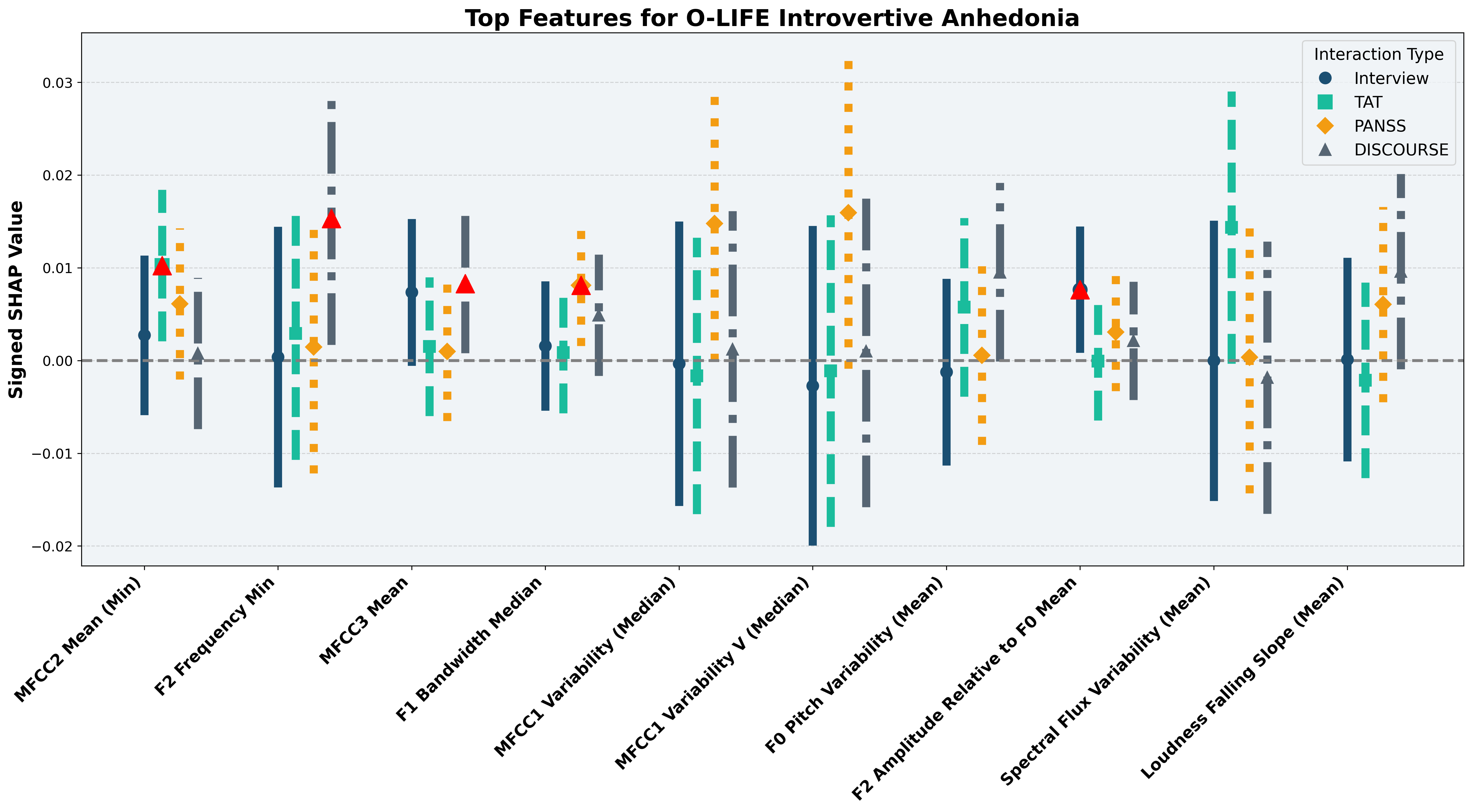}

    \vspace{0.5em}
    \includegraphics[width=0.48\textwidth]{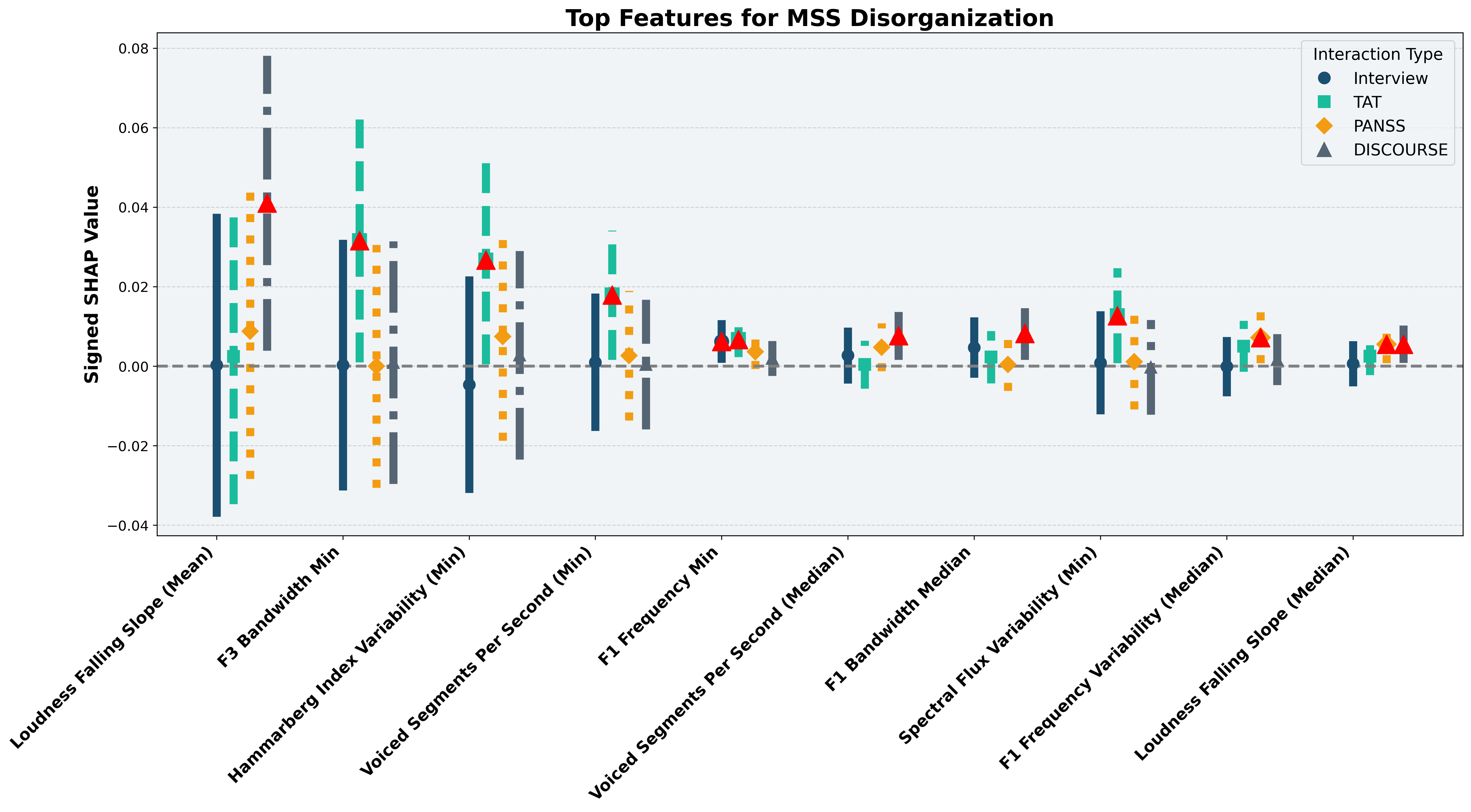}
    \hfill
    \includegraphics[width=0.48\textwidth]{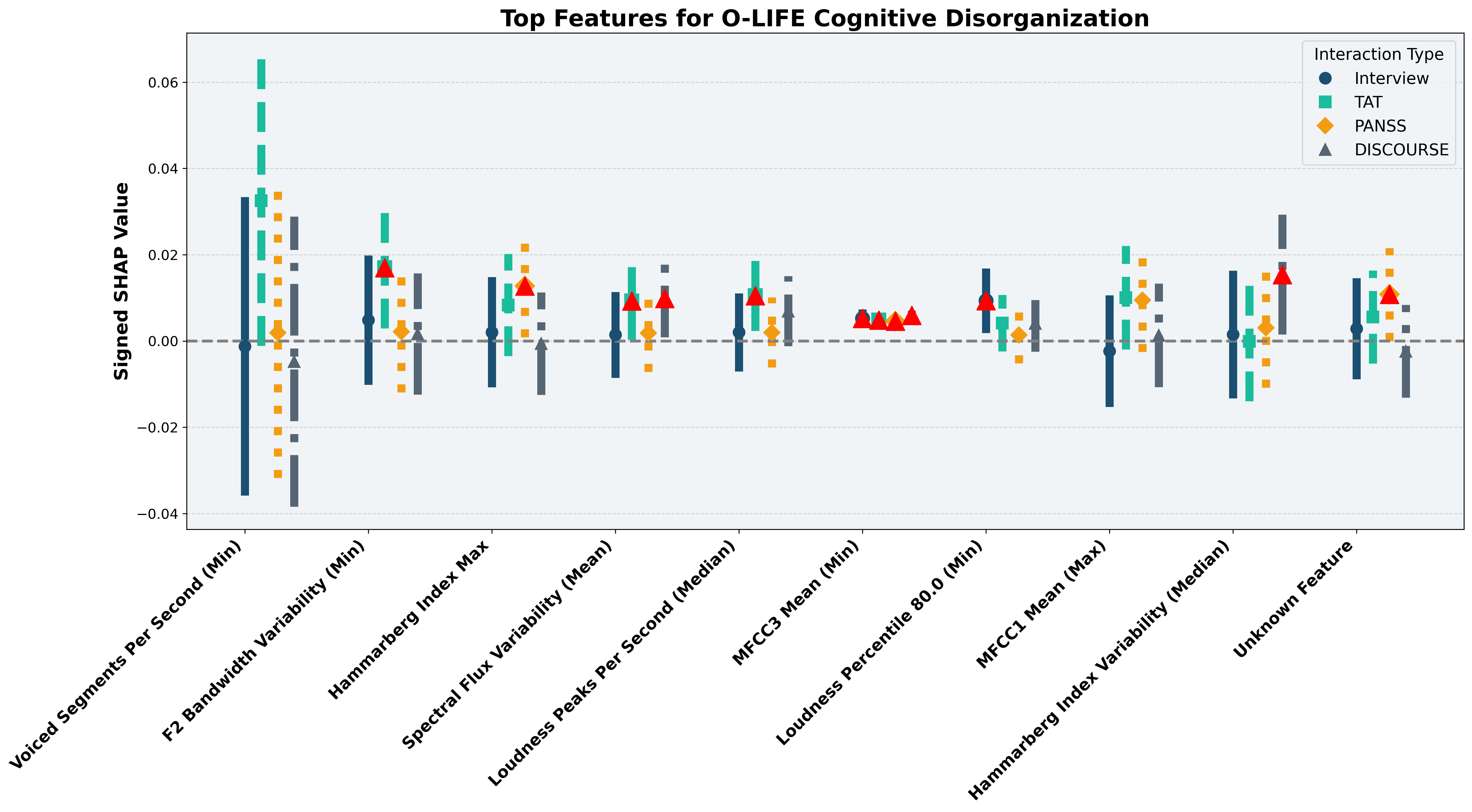}

    \vspace{0.5em}
    \includegraphics[width=0.48\textwidth]{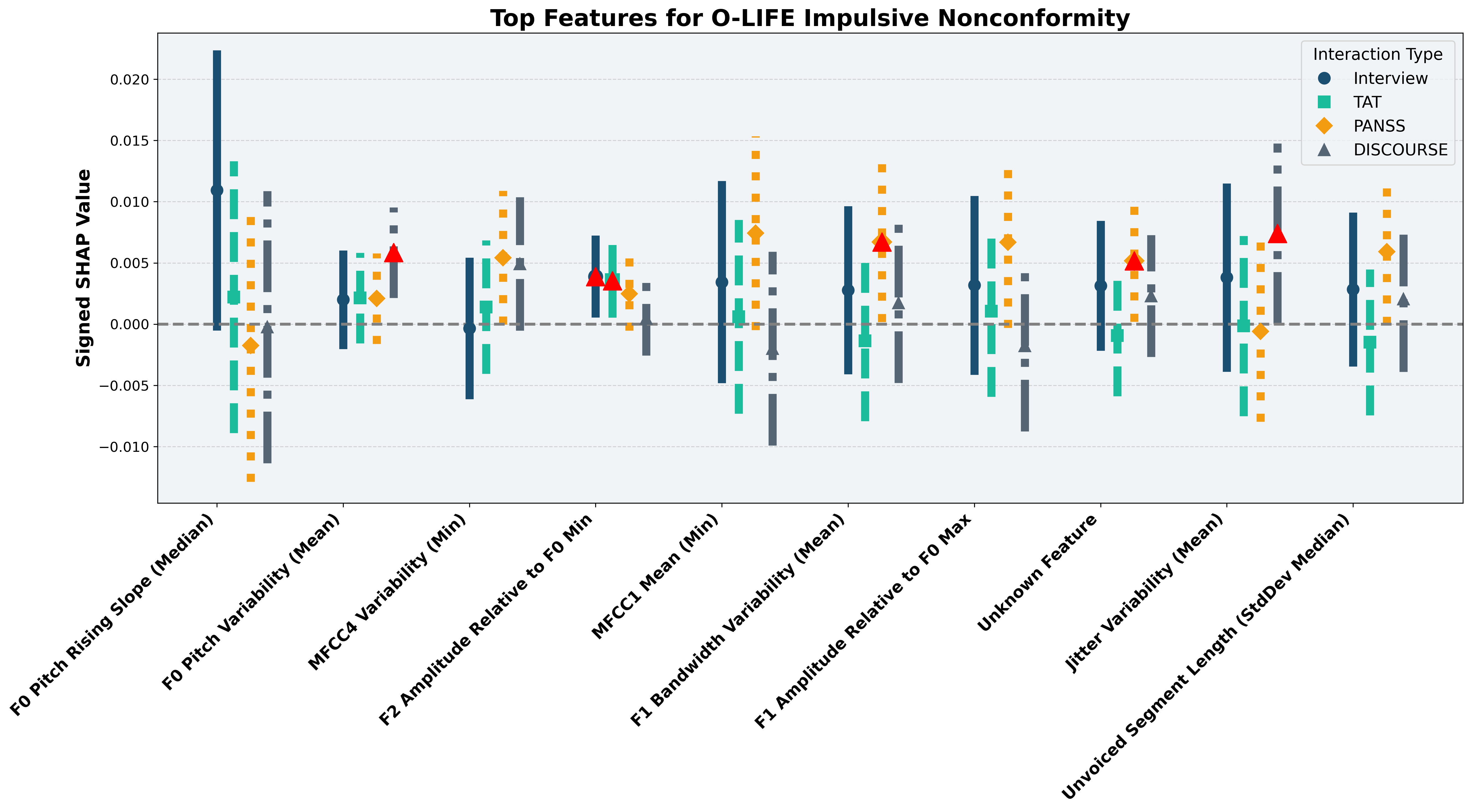}

    \caption{Top features across schizotypal traits and symptoms, organized by Positive, Negative, and Disorganized dimensions. Red triangles indicate features with a 95\% confidence interval that does not include zero.}
    \label{fig:shap_features}
\end{figure*}

\subsection{Regression results in both clinical and non-clinical settings}

Table 4 highlights the root mean square error (RMSE) results across different interaction types and models, revealing the strengths of TCF in both clinical and non-clinical settings. The distinction between PANSS and the broader trait-like measures of O-LIFE and MSS is critical to interpreting these results. This dual focus aligns with our hypothesis that speech disruptions are present across the schizophrenia spectrum, and that uncertainty-aware multimodal fusion captures these patterns more effectively than unimodal models.

In Semi-structured interactions, TCF achieves lower RMSE scores compared to GRU models, particularly in PANSS Positive and Negative symptoms (4.04 and 3.18, respectively). 
TAT tasks show a similar pattern, with TCF again producing the lowest RMSE values. The model's performance here is particularly significant for PANSS symptoms, where RMSE values for Positive and Negative symptoms drop to 3.68 and 3.45, respectively.

In DISCOURSE interactions, the RMSE values for PANSS symptoms, though slightly higher (4.31 for Positive, 4.23 for Negative), still show that the model retains its ability to detect relevant speech patterns in less focused clinical environments.

These results highlight the flexibility of TCF, that it not only predicts symptom severity in clinical populations but also captures broader schizotypal traits across different settings.

Figure \ref{fig:model_performance_weights} compares the RMSE values for speech, text, and fusion models across Semi-structured Interviews and DISCOURSE sessions, as well as the corresponding fusion weights for speech and text modalities. The performance trends highlight the advantage of the fusion model across both interaction types, particularly in more varied and less structured tasks.

In the Semi-structured Interviews, the fusion model consistently produces lower RMSE values compared to speech and text models individually. The fusion model outperforms in tasks involving more personal and reflective content, such as "Family and Relationships" and "Health and Well-being." These categories exhibit the strongest reduction in error, underscoring the model’s ability to capture both acoustic and semantic aspects that are crucial in these domains. The text model shows slightly higher RMSE in "Education and Learning" in these question where speech patterns alone are less predictive.

The DISCOURSE sessions show a similar pattern, though with more variability in model performance across different tasks. Fusion remains consistently superior in capturing nuanced disruptions, with the "Dream Reports" task showing the lowest RMSE for the fusion model. The "Storyboard Task" reflects a slight drop in performance for the text model, likely due to its narrative structure which benefits more from acoustic input. Fusion maintains its edge across all tasks, confirming the model’s adaptability in both structured and semi-structured environments.

The lower panels in Figure \ref{fig:model_performance_weights} show the relative contribution of each modality. In Semi-structured Interviews, text consistently receives higher weights, particularly in categories like "Education and Learning," where semantic content dominates. By contrast, speech is weighted higher in topics such as "Hobbies and Leisure," where prosodic features may capture emotional or cognitive disruptions.

In DISCOURSE sessions, text again dominates the weighting across tasks, particularly in "Image Description" and "Dream Reports," where narrative content is key. However, the higher weight assigned to speech in "Storyboard Task" and "Health Narratives" indicates the importance of acoustic features in these contexts, where vocal disruptions often signal cognitive shifts. The balanced interplay between modalities supports the model's robustness in handling diverse tasks, affirming the foundation that speech and linguistic features combined provide the most reliable indicators of schizophrenia-related disruptions.

Our results show that TCF’s modality-specific variance estimates offer 
an interpretable indicator of data quality or model confidence. High $\sigma^2$ 
in acoustic features can arise from background noise or unclear pronunciations; 
clinicians seeing such high uncertainty could either re-check the audio or 
request additional, clearer samples. Similarly, high text variance might imply 
the participant’s speech is fragmented or the automatic transcription is suspect. 
In real-world psychiatric evaluations, recognizing these uncertainty signals 
could prompt a clinician to administer a different speech elicitation task 
or to confirm the findings with a second observer. This fosters 
trust in automated assessments and facilitates more nuanced, patient-specific 
diagnostic decisions.

\section{Discussion}
Using speech data from both clinical and non-clinical individuals from the psychosis spectrum, this study found that incorporating uncertainty modeling into speech analysis improves the reliability of predicting psychosis-related traits. Compared to models without uncertainty estimation, our approach reduced the Expected Calibration Error (ECE) to 4.5e-2 and achieved an F1-score of 83\%, indicating better alignment between predicted confidence and actual outcomes. Participants completed structured interviews, semi-structured autobiographical tasks, and narrative-driven interactions, allowing us to analyze speech variability across different contexts. Below, we discuss our main results in more detail.

\subsection{Acoustic Patterns and Schizotypal Traits}
 Figure \ref{fig:shap_features} highlights the most predictive acoustic features for PANSS, MSS, and O-LIFE scales across various interaction types. The analysis reveals systematic speech variations tied to positive, negative, and disorganized symptomatology, showing how specific traits manifest acoustically in both clinical and subclinical populations.
 
\textbf{Positive Symptoms and Cognitive-Perceptual Traits}
Acoustic markers reflect heightened arousal, emotional intensity, and cognitive-perceptual anomalies in individuals with positive schizotypy traits. In MSS-CP, participants showed a higher mean fundamental frequency (F0), indicating elevated pitch. This suggests increased autonomic arousal, which aligns with unusual perceptual experiences. Consistent second formant (F2) bandwidths point to stable vowel articulation. This reflects focused attention on perceptual details.
For O-LIFE-UE, median F2 bandwidth and alpha ratios of unvoiced sounds emerged as significant features. Alpha ratios measure the balance of high and low-frequency energy. Increased values suggest vocal strain or tension, possibly linked to anxiety or cognitive load.
Task-based variations highlighted these patterns. During TAT tasks, participants displayed increased alpha ratios and greater median F2 bandwidths. Image-based tasks likely amplify emotional responses and cognitive load, making vocal markers more prominent. In semi-structured interviews, mean F0 remained elevated, and F2 bandwidths stayed consistent. This suggests that less structured interactions allow stronger expression of cognitive-perceptual anomalies.
In contrast, discourse tasks constrained vocal features. Structured interactions limited expressive variability, reducing pitch elevation and spectral changes.
PANSS Positive scores showed additional patterns. Higher pitch variability, spectral fluctuations, and dynamic loudness slopes dominated these profiles. These acoustic features signal emotional instability and fragmented thought processes. They were most evident during unstructured speech, where free-form expression highlights cognitive disorganization.

\textbf{Negative Schizotypy Traits and Symptoms} Negative symptoms and interpersonal traits influence speech patterns in both clinical and subclinical populations. Participants with higher PANSS Negative scores, MSS-IP traits, and subclinical O-LIFE-IA show reduced prosody and restricted vocal variability. Narrower F3 bandwidths and changes in Mel Frequency Cepstral Coefficients (MFCCs) reflect these effects.
Narrower F3 bandwidths indicate limited movement of articulators like the tongue and lips. This articulatory restriction results in less variability in speech sounds. It aligns with flattened affect, emotional disengagement, and reduced expressivity. Changes in MFCCs show spectral uniformity and reduced voice quality. These patterns suggest diminished muscular engagement and motivation during speech production.
PANSS interviews amplify negative symptoms due to their structured nature. In these settings, participants display flat prosody, reduced pitch modulation, and slow loudness slopes. Speech remains monotone with minimal spectral flux, emphasizing emotional blunting and alogia.
Semi-structured interviews and conversational tasks like TAT reduce the severity of these acoustic markers. Spontaneous engagement encourages more natural speech variability. However, prosodic flattening and reduced articulation persist, though less pronounced. DISCOURSE tasks further mask negative symptom expression. The structured and predictable format minimizes speech variability, masking acoustic signatures of emotional disengagement.
Subclinical traits like Introvertive Anhedonia share similar features. Individuals exhibit low pitch variability and uniform spectral patterns. Their monotone voice reflects blunted affect and social withdrawal. These acoustic markers highlight the overlap between clinical and subclinical negative symptoms.
Across all dimensions, negative symptoms consistently produce flattened prosody and restricted articulation. Reduced pitch modulation, spectral variability, and vocal engagement signal emotional disengagement and diminished expressivity. Interaction types and settings influence the degree of symptom expression, but universal flattening remains a core feature of negative symptoms in speech.

\textbf{Disorganized Thinking and Behavior} Participants with high O-LIFE-CD subscale scores show increased loudness variability, spectral flux, and fluency disruptions. These markers reflect fragmented cognitive processes, thought disorder, and impaired articulatory control.
Loudness variability and spectral flux highlight erratic modulation of vocal intensity and spectral energy. Participants produce fluctuating volume and inconsistent pitch patterns. Spectral flux captures rapid changes in energy across speech frequencies, while falling loudness slopes suggest inconsistent control over speech intensity.
Reduced voiced segments per second reveal fluency disruptions like hesitations, pauses, and slow speech. These disruptions could point to difficulties organizing thoughts during communication. Increased MFCC variability further emphasizes impaired articulatory control. Variability in these features reflects instability in the coordination of speech muscles, which aligns with fragmented cognitive processing.
Interaction contexts influence the severity of disorganized symptoms. During DISCOURSE tasks, structured demands amplify loudness variability, spectral flux, and fluency disruptions. Participants struggle to maintain consistent speech flow, producing abrupt loudness shifts and irregular spectral energy. Temporal and spectral instability are most pronounced in these structured settings. In semi-structured interviews and photo-elicitation tasks (TAT), these disruptions remain but appear less severe. Open-ended interactions encourage flexibility, allowing participants to navigate speech more freely.

\begin{table*}[htbp]
\centering
\rowcolors{2}{LightGrey}{white}  
\begin{tabularx}{\textwidth}{>{\raggedright\arraybackslash}X | l | c | c | c | c | c | c | c | c | c | c }
\toprule
\textbf{Interaction Type} & \textbf{Model} & \multicolumn{2}{c}{\textbf{PANSS}} & \multicolumn{3}{c}{\textbf{MSS}} & \multicolumn{4}{c}{\textbf{O-LIFE}} \\
                          &                & \textbf{Positive} & \textbf{Negative} & \textbf{CP} & \textbf{IP} & \textbf{DO} & \textbf{UE} & \textbf{IA} & \textbf{CD} & \textbf{IN} \\
\midrule
\textbf{Interview} & GRU (Speech)     & 5.12 & 4.25 & 5.50 & 4.45 & 6.12 & 7.31 & 5.12 & 6.54 & 4.04 \\
                      & GRU (Language)   & 4.78 & 3.57 & 4.87 & 4.09 & 5.14 & 7.06 & 4.81 & 6.22 & 3.19 \\
                      & TCF       & \textbf{4.04} & \textbf{3.18} & \textbf{3.64} & \textbf{3.79} & \textbf{5.16} & \textbf{5.46} & \textbf{3.68} & \textbf{4.49} & \textbf{3.08} \\
\midrule
\textbf{TAT} & GRU (Speech)     & 4.83 & 4.65 & 4.65 & 4.85 & 6.56 & 6.83 & 5.65 & 6.85 & 4.15 \\
                          & GRU (Language)   & 4.49 & 4.58 & 3.29 & 4.17 & 6.19 & 5.63 & 5.83 & 4.68 & 3.84 \\
                          & TCF       & \textbf{3.68} & \textbf{3.45} & \textbf{3.32} & \textbf{3.66} & \textbf{3.38} & \textbf{5.45} & \textbf{3.82} & \textbf{4.43} & \textbf{3.14} \\
\midrule
\textbf{PANSS} & GRU (Speech)     & 5.20 & 4.80 & 5.50 & 4.90 & 6.60 & 7.20 & 5.70 & 7.10 & 4.30 \\
               & GRU (Language)   & 4.68 & 3.57 & 4.91 & 4.14 & 5.07 & 7.21 & 4.97 & 6.12 & 3.42 \\
               & TCF       & \textbf{3.67} & \textbf{3.34} & \textbf{3.12} & \textbf{3.76} & \textbf{3.57} & \textbf{5.08} & \textbf{3.73} & \textbf{4.45} & \textbf{3.92} \\
\midrule
\textbf{DISCOURSE} & GRU (Speech)     & 5.89 & 5.68 & 5.65 & 6.00 & 7.80 & 7.92 & 6.90 & 6.80 & 5.25 \\
                  & GRU (Language)   & 4.44 & 4.47 & 4.18 & 5.08 & 7.14 & 6.41 & 6.56 & 5.48 & 4.59 \\
                  & TCF       & \textbf{4.31} & \textbf{4.23} & \textbf{4.15} & \textbf{4.75} & \textbf{4.63} & \textbf{6.01} & \textbf{4.92} & \textbf{5.61} & \textbf{4.34} \\
\bottomrule
\end{tabularx}
\caption{Root Mean Square Error (RMSE↓) results for different models and interaction types. The table groups models into Speech (GRU), Language (GRU), and Multimodal models (TCF). The interaction types include PANSS (Positive, Negative), MSS with Cognitive-Perceptual (CP), Interpersonal (IP), and Disorganization (DO), and O-LIFE with Unusual Experiences (UE), Introvertive Anhedonia (IA), Cognitive Disorganization (CD), and Impulsive Nonconformity (IN). Bold values represent the lowest RMSE in each column.}
\end{table*}

\textbf{Common Features Across Symptoms} Across schizotypal traits and symptoms, specific acoustic features consistently appear. Alterations in MFCCs and formant bandwidths (F2 and F3) emerge as significant markers across scales. These changes reflect disruptions in the spectral characteristics of speech, influenced by articulatory and phonatory modifications. Articulatory changes result from altered vocal tract movements, while phonatory changes reflect vocal fold tension and subglottal pressure. Together, these features signal underlying motor control and coordination deficits.
In addition to spectral changes, prosodic features like pitch variability and loudness slopes provide further insight. Pitch instability and spectral variability are prominent in positive and disorganized traits. In contrast, prosodic flattening and reduced articulatory dynamism mark negative symptoms. Disorganized traits also display temporal irregularities, reflecting fragmented cognitive processes.
The interaction context strongly influences these acoustic features. Open-ended tasks, such as semi-structured interviews and photo-elicitation tasks (TAT), amplify natural variability in speech. In these settings, participants show more pronounced changes in pitch, loudness, and spectral markers. This variability reveals underlying cognitive and emotional disturbances, which structured tasks may suppress. In contrast, tasks like PANSS assessments and DISCOURSE exercises constrain speech. These structured settings emphasize prosodic flattening and temporal disruptions, particularly in negative and disorganized traits.
Acoustic markers align with specific schizotypal symptom domains. Positive symptoms and cognitive-perceptual traits are characterized by pitch instability and spectral variability. These reflect cognitive and emotional disruptions. Negative symptoms and interpersonal traits show flattened prosody and reduced articulatory dynamics, signaling emotional withdrawal and motor deficits. Disorganized traits manifest through temporal disruptions and spectral irregularities, mirroring fragmented thought and cognition.

\subsection{Language Patterns and Schizotypal Traits}

\begin{figure*}[htp] 
    \centering
    \includegraphics[width=\textwidth]{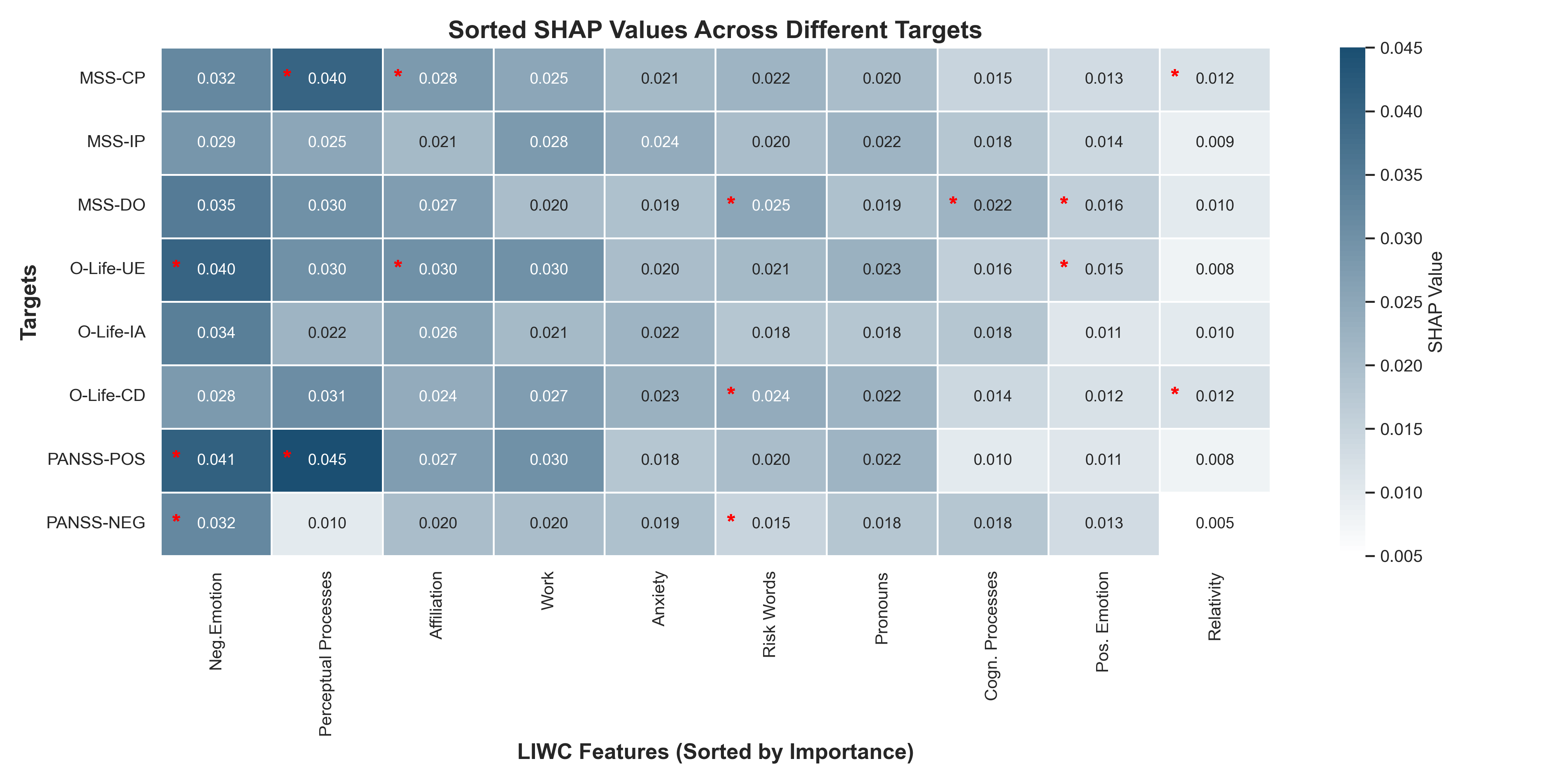}
    \caption{Heatmap of SHAP values for various LIWC features across different targets. Features are sorted by their average SHAP importance, and significant values are marked with red stars. The targets include PANSS (Positive, Negative), MSS with Cognitive-Perceptual (CP), Interpersonal (IP), and Disorganization (DO), and O-LIFE with Unusual Experiences (UE), Introvertive Anhedonia (IA), Cognitive Disorganization (CD), and Impulsive Nonconformity (IN).}
    \label{fig:shap_values_heatmap}
\end{figure*}

Figure \ref{fig:shap_values_heatmap} highlights the key linguistic patterns associated with schizotypal traits and symptoms across PANSS, MSS, and O-LIFE scales. 

\textbf{Positive Traits and Symptoms}
In MSS-CP and O-LIFE-UE, participants used perceptual process words, affiliation terms, and emotional language. Perceptual process words, such as those related to sight, sound, or sensation, reflect increased sensory awareness and altered perceptions common in positive schizotypy \cite{asimakidou2022positive}. Affiliation terms point to a focus on social connections, suggesting that unusual perceptions often involve interpersonal themes. The co-occurrence of negative and positive emotion words in O-LIFE-UE could show emotional complexity, reflecting distress mixed with excitement or arousal.
For PANSS Positive, perceptual process words and negative emotion terms dominate. The prominence of perceptual words aligns with clinically significant hallucinations and delusions, where sensory distortions are core features \cite{buck2015lexical}. Negative emotion words suggest emotional distress often tied to these experiences. Interestingly, the SHAP values for affiliation terms are less prominent in PANSS, likely reflecting reduced interpersonal coherence during clinical positive symptoms. Across both clinical and non-clinical groups, the prominence of perceptual language connects heightened sensory awareness in schizotypy with more severe perceptual distortions in psychosis. Emotional markers reveal distress as a shared thread, while social language appears more relevant to subclinical cognitive-perceptual traits.

\textbf{Negative Traits and Symptoms}
In PANSS Negative scores, participants used more negative emotion and risk words. Despite the characteristic emotional flatness of negative symptoms, negative emotion words indicate underlying emotional struggle or dissatisfaction, even when outward expression is limited. Risk words reflect cognitive disconnection, possibly signaling unconventional or fragmented thought processes \cite{buck2015lexical}.
In MSS-IP and O-LIFE-IE, participants used fewer social words and positive emotion terms. Affiliation terms, which appeared in positive traits, are significantly reduced here, highlighting social disengagement and emotional withdrawal. The lack of emotional diversity aligns with anhedonia and poverty of thought, core features of negative traits \cite{horan2006anhedonia}.
These patterns show a clear link between clinical negative symptoms and subclinical interpersonal deficits. While PANSS Negative reflects pronounced emotional flatness and cognitive poverty, MSS and O-LIFE reveal milder but consistent reductions in social engagement and emotional expressivity.

\textbf{Disorganized Traits and Symptoms}
In MSS-DO and O-LIFE-CD, participants used more risk words, cognitive process words, and relativity terms. Risk words suggest impulsivity and unconventional speech patterns, consistent with cognitive fragmentation in disorganized traits. The use of cognitive process words reflects mental effort; however, in individuals exhibiting disorganized symptoms, their speech may still lack coherence due to underlying thought disturbances \cite{chang2022language}. Risk words and relativity terms also mark disorganized traits. This suggests that cognitive fragmentation, impulsivity, and challenges with abstract processing are shared linguistic markers across clinical and non-clinical disorganization.

\textbf{Common Linguistic Patterns and Contextual Influence}
Across schizotypy dimensions, shared linguistic markers include negative emotion words, perceptual process terms, and risk words. Perceptual terms dominate positive traits, revealing a connection between altered perceptions and heightened sensory awareness. Negative emotion words appear across both positive and negative symptoms. Risk words reflect cognitive disorganization, impulsivity, and unconventional thinking, aligning closely with disorganized traits.
The influence of interaction context is notable. Structured tasks, like PANSS interviews, bolden negative symptoms by constraining speech, leading to reduced emotional expression and cognitive complexity. Open-ended tasks, like TAT narratives, highlight disorganized and cognitive-perceptual traits, as participants produce richer, more expressive language that exposes subtle thought disruptions.

Positive symptoms and traits consistently feature perceptual language and emotional intensity, reflecting altered awareness and hyperarousal. Negative symptoms traits show reduced emotional diversity and fewer social words, aligning with emotional flatness and social disengagement. Disorganized traits manifest through cognitive process words, risk terms, and relativity words, signaling fragmented thought and linguistic incoherence. Task structure plays a critical role, with open-ended interactions amplifying disorganized traits and structured settings constraining emotional variability.

\subsection{Psychosis Heterogeneity and the Importance of Modeling Uncertainty}
Psychosis spans a spectrum of presentations, from subtle schizotypal traits to severe symptomatology in schizophrenia. Our uncertainty-aware TCF model mitigates the risk of overconfidence across this varied landscape. For instance, 
participants with mild disorganized speech (subclinical schizotypy) often show small yet meaningful cues that naive approaches may overlook. By assigning higher variance to uncertain segments, TCF flags borderline cases that merit 
additional scrutiny rather than a hasty diagnosis. This distinction is especially important when moving from subclinical to clinical thresholds, where misclassification carries serious consequences.

We also observed that structured tasks (e.g., PANSS interviews) typically generate more uniform speech, often accentuating negative symptoms. In contrast, open-ended interactions (e.g., autobiographical interviews) can elevate disorganized or 
positive traits through tangential or imaginative speech. This context-driven variability underscores the need to model heteroscedastic noise. TCF responds to these fluctuations by increasing the variance estimate when speech coherence drops, 
thus reducing the risk of overconfident predictions. These findings support our goal of identifying psychosis-related speech disruptions that emerge in unpredictable, variable conditions. They also highlight how TCF effectively adapts to these contextual shifts using modality-specific uncertainty estimates.

\subsection{Limitations}
Although speech analysis offers a systematic and quantifiable approach compared to subjective clinical assessments, interpretative biases remain unavoidable. Variability in speech production across tasks, linguistic backgrounds, and recording conditions challenges generalizability. Our model addresses these uncertainties, but standardizing speech-based assessments and adapting them to diverse populations remains crucial. Refining these tools will improve their clinical applicability, enhance early detection, and support more reliable symptom monitoring and treatment planning across the psychosis spectrum. Addressing these challenges will strengthen the role of speech-based modeling in psychiatry, enabling more objective and personalized assessments. Future research should prioritize longitudinal studies and real-world applications to refine these methods and improve clinical decision-making.

\section{Conclusion}
Our study showed that speech disruptions across the psychosis spectrum are measurable, consistent, and context-dependent. By integrating acoustic and linguistic features through an uncertainty-aware model, we captured subtle patterns linked to positive, negative, and disorganized traits. Positive symptoms and cognitive-perceptual traits exhibited enhanced sensory awareness, reflected in increased pitch variability and perceptual language. Negative symptoms showed reduced vocal variability, flattened prosody, and diminished emotional expression. Disorganized traits manifested as erratic spectral changes, loudness instability, and fragmented fluency. These patterns appear consistently across clinical (PANSS) and subclinical (MSS, O-LIFE) measures, supporting the psychosis continuum hypothesis. The interaction context played a critical role in symptom expression. Structured tasks, such as PANSS interviews, emphasized negative traits, while open-ended tasks amplified disorganized and positive features. This highlights the importance of task variability in clinical assessments. Our uncertainty-aware fusion model improved prediction accuracy and reliability by addressing speech variability across individuals and settings. It provides a robust tool for identifying speech markers of psychosis and schizotypy, offering new pathways for objective assessment.         

\section*{Data and code availability}
Data and code used to support these findings are available from the corresponding author upon reasonable request.

\section*{Acknowledgements}
We are grateful to all participants for their contributions. We also thank Anna Steiner, Linus Hany, and Ueli Stocker for their help with data collection.

\section*{Conflicts of interest} PH has received grants and honoraria from Novartis, Lundbeck, Takeda, Mepha, Janssen, Boehringer Ingelheim, Neurolite and OM Pharma outside of this work. No other conflicts of interest were reported.

\section*{Funding} This work was supported by the European Union (GA 101080251 - TRUSTING) and by the Swiss National Science Foundation (POZHP1\_191938/1).

\printbibliography 
\newpage  
\appendix

\onecolumn
\section{Appendix: Supplementary Material}

\subsection{Kernel Density Estimation (KDE) Plots for Psychological Measures}

\begin{figure*}[ht!]
    \centering
    \includegraphics[width=0.9\textwidth]{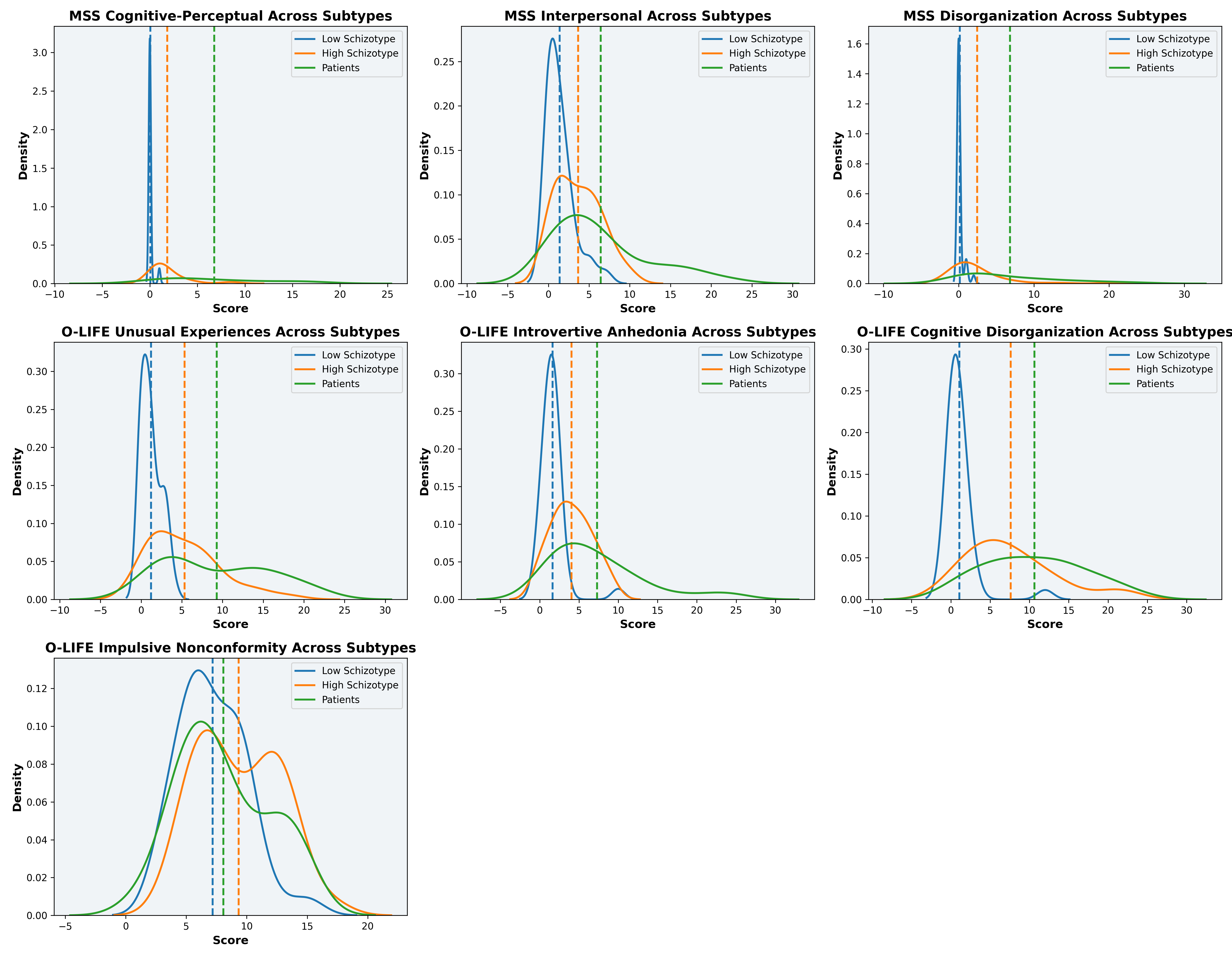}
    \caption{Kernel Density Estimation (KDE) plots with mean lines for different psychological measures across subtypes. Each subplot represents a distinct MSS or O-LIFE subscale. The three groups displayed are low schizotypy, high schizotypy, and patients. Dashed vertical lines indicate the mean values for each group within each subplot. The number of participants in each group is as follows: low schizotypy ($n = 45$), high schizotypy ($n = 37$), and patients ($n = 32$).}
    \label{fig:mss_olife_subtypes}
\end{figure*}

\newpage  

\subsection{Standard Deviations of F1 Scores for Speech and Language Models}

\begin{table*}[ht!]
\centering
\rowcolors{2}{LightGrey}{white}  
\begin{tabularx}{\textwidth}{lX|c>{\columncolor{LightBlue}}c|c>{\columncolor{LightBlue}}c|c>{\columncolor{LightBlue}}c|c>{\columncolor{LightBlue}}c}
\toprule
& \textbf{Interaction Type} & \multicolumn{1}{c}{\textbf{Interview}} & \multicolumn{1}{c}{\textbf{TAT}} & \multicolumn{1}{c}{\textbf{PANSS}} & \multicolumn{1}{c}{\textbf{DISCOURSE}} \\
& \textbf{Metric} & \textbf{STD (F1)} & \textbf{STD (F1)} & \textbf{STD (F1)} & \textbf{STD (F1)} \\
\midrule
\textbf{Speech} & Random Forest & 2.5 & 3.0 & 2.8 & 3.2 \\
& LDA & 3.0 & 3.1 & 3.4 & 3.2 \\
& SVM & 2.4 & 2.5 & 2.2 & 2.3 \\
& GRU & 1.8 & 2.0 & 1.9 & 2.1 \\
& DEEPSPECTRUM & 1.7 & 1.7 & 1.8 & 1.9 \\
& wav2vec-Large & 1.7 & 1.8 & 1.6 & 1.7 \\
\midrule
\textbf{Language} & Random Forest & 2.5 & 3.0 & 2.8 & 3.2 \\
& LDA & 3.1 & 3.0 & 3.3 & 3.6 \\
& SVM & 2.5 & 3.0 & 2.7 & 2.6 \\
& GRU & 1.8 & 2.0 & 1.8 & 2.1 \\
& Fine-tuned BERT & 1.5 & 1.6 & 1.8 & 1.6 \\
& Fine-tuned RoBERTa & 1.5 & 1.7 & 1.3 & 1.4 \\
\bottomrule
\end{tabularx}
\caption{Standard deviations (STD) of F1 Scores for Speech and Language Models across different interaction types: Semi-structured Autobiographical Interview (Interview), Thematic Apperception Test (TAT), The PANSS Clinical Interview (PANSS), and DISCOURSE session.}
\end{table*}
\begin{table*}[htbp]
\centering
\rowcolors{2}{LightGrey}{white}
\begin{tabular}{l|c|c|c|c|c}
\toprule
\textbf{Models} & \textbf{Interview} & \textbf{TAT} & \textbf{PANSS} & \textbf{DISCOURSE} \\
                & \textbf{STD (F1)}  & \textbf{STD (F1)}  & \textbf{STD (F1)}  & \textbf{STD (F1)}  \\
\midrule
Early Fusion    & 0.02 & 0.02 & 0.03 & 0.03 \\
Late Fusion     & 0.01 & 0.02 & 0.02 & 0.03 \\
Context Fusion  & 0.01 & 0.02 & 0.02 & 0.02 \\
TCF             & 0.01 & 0.01 & 0.02 & 0.02 \\
\bottomrule
\end{tabular}
\caption{Standard deviations (STD) of F1 Scores for Multimodal Fusion Models across different interaction types: Interview, Thematic Apperception Test (TAT), PANSS, and DISCOURSE.}
\end{table*}

\subsection{Mean Absolute Error (MAE) Results for Different Models}

\begin{table*}[htbp]
\centering
\rowcolors{2}{LightGrey}{white}  
\begin{tabularx}{\textwidth}{>{\raggedright\arraybackslash}X | l | c | c | c | c | c | c | c | c | c | c }
\toprule
\textbf{Interaction Type} & \textbf{Model} & \multicolumn{2}{c}{\textbf{PANSS}} & \multicolumn{3}{c}{\textbf{MSS}} & \multicolumn{4}{c}{\textbf{O-LIFE}} \\
                          &                & \textbf{Positive} & \textbf{Negative} & \textbf{CP} & \textbf{IP} & \textbf{DO} & \textbf{UE} & \textbf{IA} & \textbf{CD} & \textbf{IN} \\
\midrule
\textbf{Interview} & GRU (Speech)     & 3.50 & 3.10 & 3.95 & 3.40 & 4.75 & 5.60 & 3.50 & 4.85 & 2.35 \\
                      & GRU (Language)   & 2.45 & 2.90 & 3.85 & 3.10 & 4.05 & 4.65 & 3.20 & 4.35 & 2.15 \\
                      & TCF       & \textbf{2.30} & \textbf{2.50} & \textbf{3.00} & \textbf{2.80} & \textbf{3.25} & \textbf{4.20} & \textbf{2.10} & \textbf{3.85} & \textbf{2.00} \\
\midrule
\textbf{TAT} & GRU (Speech)     & 3.20 & 3.30 & 4.00 & 3.25 & 5.20 & 5.40 & 3.30 & 5.10 & 2.50 \\
                          & GRU (Language)   & 3.00 & 3.00 & 3.50 & 3.00 & 4.10 & 4.50 & 3.00 & 4.20 & 2.40 \\
                          & TCF       & \textbf{2.75} & \textbf{2.40} & \textbf{2.90} & \textbf{2.70} & \textbf{3.15} & \textbf{4.10} & \textbf{2.00} & \textbf{3.75} & \textbf{1.90} \\
\midrule
\textbf{PANSS} & GRU (Speech)     & 3.00 & 2.85 & 3.20 & 3.05 & 4.00 & 4.90 & 2.80 & 4.40 & 2.30 \\
               & GRU (Language)   & 2.80 & 2.75 & 3.10 & 2.90 & 3.80 & 4.50 & 2.60 & 4.20 & 2.20 \\
               & TCF       & \textbf{2.50} & \textbf{2.30} & \textbf{2.75} & \textbf{2.60} & \textbf{3.10} & \textbf{3.90} & \textbf{1.90} & \textbf{3.50} & \textbf{1.80} \\
\midrule
\textbf{DISCOURSE} & GRU (Speech)     & 3.60 & 3.40 & 4.10 & 3.50 & 5.50 & 5.80 & 3.70 & 5.30 & 2.65 \\
                  & GRU (Language)   & 3.40 & 3.20 & 3.70 & 3.20 & 4.30 & 4.80 & 3.30 & 4.50 & 2.50 \\
                  & TCF       & \textbf{2.85} & \textbf{2.60} & \textbf{3.20} & \textbf{2.90} & \textbf{3.50} & \textbf{4.00} & \textbf{2.10} & \textbf{3.60} & \textbf{2.00} \\
\bottomrule
\end{tabularx}
\caption{Mean Absolute Error (MAE↓) results for different models and interaction types. The table groups models into Speech (GRU), Language (GRU), and Multimodal models (TCF). The interaction types include PANSS (Positive, Negative), MSS with Cognitive-Perceptual (CP), Interpersonal (IP), and Disorganization (DO), and O-LIFE with Unusual Experiences (UE), Introvertive Anhedonia (IA), Cognitive Disorganization (CD), and Impulsive Nonconformity (IN). Bold values represent the lowest MAE in each column.}
\end{table*}

\end{document}